\documentclass{article}
\usepackage{2026arXiv}

\usepackage{times}
\usepackage{soul}
\usepackage{url}
\usepackage[hidelinks]{hyperref}
\usepackage[utf8]{inputenc}
\usepackage[small]{caption}
\usepackage{graphicx}
\usepackage{amsmath}
\usepackage{amsthm}
\usepackage{amsmath}
\usepackage[capitalise]{cleveref}
\usepackage{booktabs}
\usepackage{algorithm}
\usepackage{algorithmic}
\usepackage[switch]{lineno}
\usepackage{amsfonts}
\usepackage{subcaption}

\usepackage{xcolor}
\usepackage{pifont}
\newcommand{\cmark}{\ding{51}}
\newcommand{\xmark}{\textcolor{gray!40}{\ding{55}}}
\usepackage{multirow}
\usepackage{enumitem}
\usepackage{siunitx}
\usepackage[normalem]{ulem}

\title{Physically Guided Visual Mass Estimation from a Single RGB Image}

\author{
Sungjae Lee$^1$
\and
Junhan Jeong$^1$\and
Yeonjoo Hong$^1$\And
Kwang In Kim$^{1,2}$\\
\affiliations
$^1$Graduate School of Artificial Intelligence, POSTECH, South Korea\\
$^2$Department of Electrical Engineering, POSTECH, South Korea\\
\emails
\{leeeesj, junhanjeong, yeonjooh, kwangin.kim\}@postech.ac.kr
}

\begin{document}
\maketitle

\begin{abstract}
Estimating object mass from visual input is challenging because mass depends jointly on geometric volume and material-dependent density, neither of which is directly observable from RGB appearance. Consequently, mass prediction from pixels is ill-posed and therefore benefits from physically meaningful representations to constrain the space of plausible solutions. We propose a physically structured framework for single-image mass estimation that addresses this ambiguity by aligning visual cues with the physical factors governing mass. From a single RGB image, we recover object-centric three-dimensional geometry via monocular depth estimation to inform volume and extract coarse material semantics using a vision-language model to guide density-related reasoning. These geometry, semantic, and appearance representations are fused through an instance-adaptive gating mechanism, and two physically guided latent factors (volume- and density-related) are predicted through separate regression heads under mass-only supervision. Experiments on \emph{image2mass} and \emph{ABO-500} show that the proposed method consistently outperforms state-of-the-art methods.
\end{abstract}

\section{Introduction}
Robotic manipulation demands more than geometric perception. It requires an understanding of the physical principles that govern object interactions. Beyond shape and pose, physical properties such as mass, stiffness, and friction directly influence inertial effects, contact stability, and common failure modes. Accordingly, recent work has emphasized physically grounded reasoning, from detecting violations of real-world dynamics~\cite{LGC25} to learning force-related intuition through human contact~\cite{ACL25}. Together, these efforts point to a central challenge: predicting physically relevant object attributes \emph{before} interaction to enable safer and more robust manipulation.

Among these physical properties, \emph{mass} plays a central role. Knowledge of an object's weight enables a system to regulate grasp forces, limit excessive torques, and avoid damage to fragile objects or mechanical components. Traditionally, mass is estimated through physical interaction using force and torque sensing, for example via a small test lift~\cite{HLC20}. While effective, such contact-based approaches delay estimation until interaction has begun and may expose both the system and the object to unsafe forces, while also requiring specialized hardware or repeated exploratory actions that limit their practicality.

To address these challenges, prior work has explored non-contact methods that infer object mass directly from visual input. Existing approaches either regress mass end-to-end from RGB images using large pre-trained models~\cite{TWQ25}, or introduce intermediate representations such as thickness masks and hand-crafted geometric features~\cite{SSC17}. Some methods explicitly decompose mass into volume and density estimates, reflecting the underlying physical relationship. However, in most existing pipelines, these factors remain tightly entangled within appearance-based RGB representations, making it difficult to isolate the respective contributions of geometry and material. As a result, predictions are often hard to interpret and susceptible to systematic biases when visual cues are ambiguous (e.g., visually similar objects made of different materials).

This limitation arises not only from single-view ambiguity, but also from how visual evidence is organized for physical inference. Because mass depends jointly on geometry and material, neither of which is explicitly represented in RGB, direct mass regression in pixel space is inherently ill-posed. Robust visual mass estimation therefore benefits from representations that make geometric structure and material-related cues more explicit, instead of relying solely on appearance correlations. While retrieval-based approaches can incorporate external metadata, they often fail in realistic settings where objects are novel, unmarked, or modified and identifiers may be unavailable.

Motivated by this perspective, we adopt a physically grounded formulation where mass is expressed as the product of two complementary latent factors: a geometry-related factor and a material-related factor. Rather than entangling these factors within a single appearance-driven predictor, we use geometric and semantic cues to \emph{physically guide} their inference. Specifically, 3D geometry inferred via monocular depth estimation is used to inform volume estimation, while semantic material understanding extracted from a vision--language model (VLM) serves as a proxy for density-related reasoning. The resulting image, point cloud, and textual representations are fused using an instance-adaptive gated mechanism that learns the relative importance of each cue for a given object. This design reflects the fact that some objects are more geometry-dominated, whereas others are more strongly influenced by material-dependent density.

Experiments on the \emph{image2mass}~\cite{SSC17} and \emph{ABO-500}~\cite{ZSC24} datasets show that our single-image method significantly outperforms existing image-only and depth-augmented baselines.

\section{Related Work}
\label{s:related}
Recent work has increasingly incorporated physical reasoning into visual understanding beyond purely geometric or kinematic representations, including dynamics violation detection~\cite{LGC25} and contact-based learning of force-related cues~\cite{ACL25}. More recently, VLMs have enabled high-level reasoning about material properties, surface characteristics, and physical feasibility. For example, \emph{GraspCoT}~\cite{CDY25} employs VLM-driven inference of material and surface properties to support adaptive grasping under language instructions, while \emph{PhysVLM}~\cite{ZTZ25} models physical reachability to identify unreachable targets and predict task failure. \cite{GRD24} construct tree-structured scene representations and infers object attributes, including material and mass-related cues, to support context-aware decision making. In this work, we focus on object mass as a concrete physical attribute and study how it can be inferred from visual input in a physically interpretable manner.

\paragraph{Single-image visual mass estimation.}
Estimating object mass from a single RGB image has been studied as a non-contact alternative to force- or interaction-based methods. \emph{Image2mass}~\cite{SSC17} introduced this problem by decomposing mass estimation into intermediate geometric and material-related predictions, such as thickness and density, which are then combined to obtain mass. Subsequent work has explored end-to-end regression of mass directly from images, including approaches based on large pre-trained visual backbones~\cite{TWQ25}. While these methods demonstrate that mass can be predicted from appearance to some extent, they typically infer geometric and material cues implicitly within a single latent representation. This entanglement limits interpretability and can lead to systematic errors when visual appearance is ambiguous, for example for objects with similar shapes but different material compositions. Our approach differs in that it explicitly associates distinct sources of visual evidence with geometry-related and material-related inference, rather than entangling them within a single appearance-driven pipeline.

\paragraph{Geometry-aware representations for physical inference.}
A natural way to reduce ambiguity in single-image mass estimation is to introduce explicit geometric structure. Several works have shown that geometry-aware representations improve physical inference from visual input. In particular, depth maps, voxel grids, and other volumetric representations have been used to refine thickness or volume estimation and improve robustness under real-world imaging conditions~\cite{AM23,CM25}, demonstrating clear advantages over RGB features alone. Our method follows this perspective by using monocular depth estimation to obtain object-centric 3D geometry, which is explicitly aligned with geometry-related inference rather than treated as an auxiliary cue.

\paragraph{Semantic priors and VLMs.}
Beyond geometry, semantic understanding provides complementary cues for physical reasoning, particularly for properties governed by material composition. Recent advances in VLMs have shown that large-scale image--text pre-training encodes broad commonsense knowledge about materials and physical attributes. Several works exploit this capability to associate visual observations with physically meaningful concepts without requiring paired supervision for numerical physical quantities~\cite{ZSC24,SYC25}. However, directly prompting LLMs or VLMs to output numerical physical parameters is often unreliable and susceptible to hallucination~\cite{LMF24,CXK24,SLY25}. In contrast, we use VLMs only to extract coarse semantic descriptions of material composition, which serve as informative priors rather than direct numeric estimates. These semantic cues are embedded and fused with visual and geometric representations under mass-only supervision.

\begin{figure*}[hbt!]
\centering
\includegraphics[width=\textwidth]{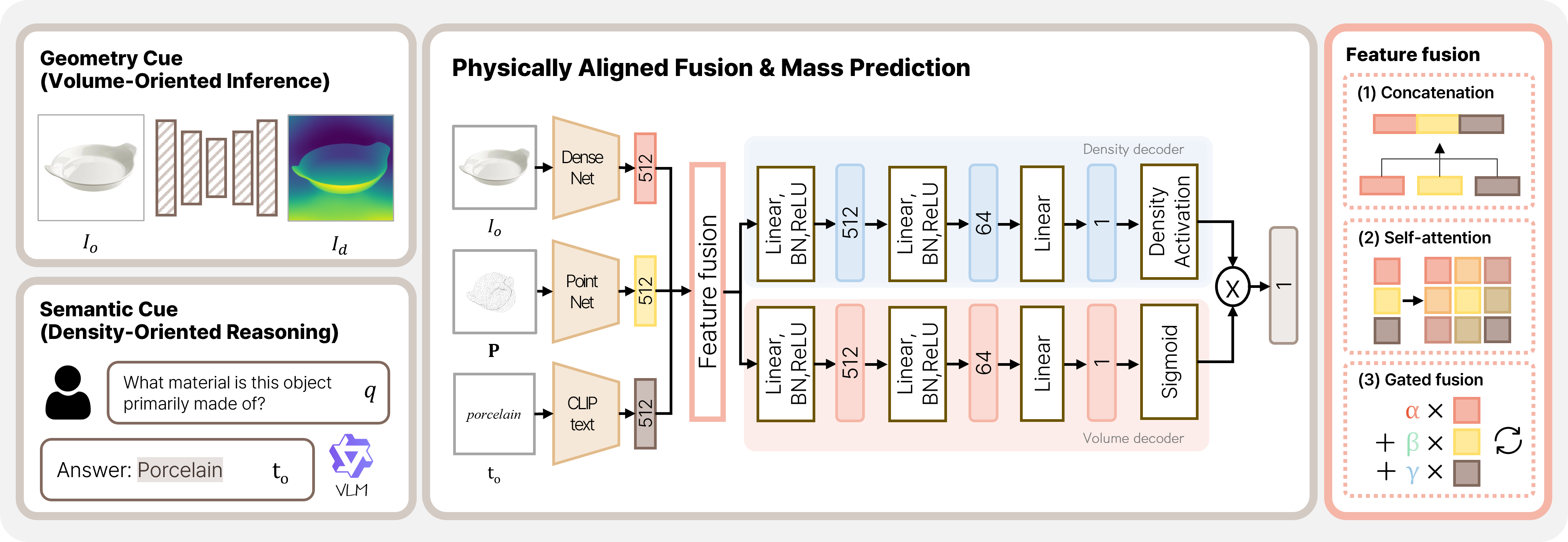}
\caption{Overview of the proposed physically structured visual mass estimation framework.
From a single RGB image, the model infers appearance features, object-centric geometry via monocular depth estimation, and a coarse semantic description of material composition. Geometry and semantic cues are explicitly aligned with volume- and density-related factor inference, respectively, while appearance provides complementary context. These representations are fused using an instance-adaptive mechanism to predict object mass. The framework is trained using mass supervision only and avoids the need for explicit density or volume labels.}
\label{f:mass}
\end{figure*}

\section{Physics-Guided Visual Mass Estimation}
\label{s:method}

\subsection{Problem Formulation and Core Principle}
Given an RGB image $I_o$ depicting an object $o$, our goal is to estimate its mass $m$. A physically grounded starting point is the relation
\begin{align}
m = V \cdot \rho,
\label{eq:mass_factor}
\end{align}
where $V$ denotes object volume and $\rho$ denotes material-dependent density. Since our model is trained with \emph{mass supervision only}, the predicted quantities $\widehat{V}$ and $\hat{\rho}$ should be interpreted as \emph{physically guided latent factors} rather than metrically calibrated estimates of true volume and density, as any rescaling $(\alpha \widehat{V}, \hat{\rho}/\alpha)$ yields the same mass. Moreover, exploiting this decomposition in a single-view setting is inherently challenging: volume depends on 3D geometry and scale, which are underconstrained in monocular images, while density depends on material composition, which is often visually ambiguous and requires semantic reasoning. As a result, multiple $(V,\rho)$ combinations can explain the same observation, making direct mass prediction ill-posed.

Our central idea is to resolve this ambiguity by explicitly aligning each latent physical factor with a source of visual evidence that is well suited to inform it. We associate the \emph{volume-related factor} with 3D geometric cues recovered via monocular depth estimation, and density-related reasoning with semantic material information inferred by a vision--language model (VLM). Rather than treating these cues as interchangeable inputs, we enforce a structured inductive bias where geometry primarily constrains volume, semantics constrains density, and appearance provides complementary context. The interaction between these factors is resolved through instance-adaptive fusion, allowing the model to emphasize different cues depending on the object.

This physically structured design yields an interpretable mass estimation framework that reflects the underlying physics in a single-image setting.

\subsection{Overview of the Multi-Cue Inference Pipeline}
The proposed framework instantiates the above principle using three complementary cues extracted from a single RGB image $I_o$: an appearance feature, an explicit geometric representation $\mathbf{P}\in\mathbb{R}^{N\times 3}$ obtained by lifting monocular depth into a point cloud, and a semantic representation $t_o$ describing the dominant material. These cues are fused into a joint embedding to predict a volume-related factor $\widehat{V}$ and a density-related factor $\hat{\rho}$, yielding the final mass estimate $\widehat{m}=\widehat{V}\hat{\rho}$. An overview is provided in \Cref{f:mass}.

\subsection{Geometry Cue for Volume-Oriented Inference}
\label{ss:depth}
Accurate volume-oriented inference benefits from access to 3D geometric structure, which is not explicitly encoded in RGB appearance alone. While an RGB image captures rich information about texture, color, and shading, these cues are only weakly correlated with absolute object size and are often entangled with lighting conditions, viewpoint, and surface reflectance. As a result, objects with similar two-dimensional projections or visual appearance can differ substantially in physical size, leading to large errors in volume and mass estimation when relying directly on appearance cues.

Importantly, this ambiguity arises from inherent single-view uncertainty and from how visual evidence is \emph{represented} for physical inference. Although all cues in our framework are ultimately inferred from a single RGB image, the representation used to organize visual evidence strongly influences which physical properties can be recovered reliably. Monocular depth estimation mitigates this issue by reorganizing appearance cues into an explicit geometric form, enforcing consistency in surface ordering and object extent along the camera ray. This 3D representation captures object shape and relative spatial extent more directly, making it better suited for volume-oriented reasoning.

While depth cameras can provide direct geometric measurements, their deployment is often constrained in real-world settings due to increased hardware cost and failures on reflective or transparent surfaces. Accordingly, we assume the availability of RGB sensors only and rely on monocular depth estimation to recover object-centric geometry.

Specifically, we predict a pixel-wise depth map $I_{\text{d}}$ from the input image $I_o$ following~\cite{CM25}. We fine-tune the \emph{GLPDepth} model~\cite{KKA22} on the ShapeNetSem dataset~\cite{CFG15} to better align the depth estimator with isolated object instances. This step is necessary because off-the-shelf monocular depth models are typically trained on 
indoor or outdoor datasets rather than object-centric imagery~\cite{YKH24,KOH24}, and consequently often exhibit imprecise object boundaries, background leakage, or inconsistent global depth scales when applied to single-object images. Since ShapeNetSem provides object-level samples with ground-truth depth, fine-tuning on this dataset allows the depth estimator to better match the data distribution encountered in the downstream mass estimation task.

To improve cross-instance scale consistency, the predicted depth values are normalized by the diagonal length of the object's bounding box. The normalized depth map is then converted into a point cloud $\mathbf{P} \in \mathbb{R}^{N \times 3}$, which serves as an explicit geometric representation for subsequent volume-oriented inference and mass estimation.

\subsection{Semantic Cue for Density-Oriented Reasoning}
\label{ss:material}
Accurate density-related inference depends primarily on an object's material composition rather than its geometric shape. Objects with similar volumes can differ substantially in mass due to differences in material (e.g., plastic versus metal) making density a critical factor for mass prediction. However, unlike geometry, density is not directly observable from spatial structure and cannot be recovered reliably from pixel-level appearance alone.

As with volume estimation, this challenge arises from representational limitations rather than missing information. Visual appearance provides indirect cues about material, such as texture and surface finish, but these cues are often ambiguous and shared across materials with vastly different densities. Learning a direct mapping from RGB appearance to numerical density is therefore difficult, particularly in the absence of large-scale, accurately annotated datasets. Indeed, obtaining paired visual observations and ground-truth physical properties such as density is costly and rarely feasible at scale, as it requires manual measurement or specialized instrumentation.

Recent work has shown that VLMs encode broad commonsense knowledge about the physical world acquired through large-scale image-text pre-training. Representative examples such as \emph{NeRF2Physics}~\cite{ZSC24} and \emph{PUGS}~\cite{SYC25} demonstrate that VLMs can associate visual appearance with physically relevant concepts without relying on explicitly paired image--property supervision. This capability provides an opportunity to incorporate high-level physical reasoning into visual inference pipelines without requiring direct numerical prediction of physical quantities.

We query the \emph{Qwen2.5-VL} model~\cite{YYZ24} with the prompt $q$ \textit{``What is this object primarily made of?''} to obtain a coarse semantic description $t_o = \mathrm{VLM}(I_o, q)$ of the object’s material composition, which provides a weak but informative cue for density-related inference.

Crucially, we do not interpret this textual output as a direct or precise estimate of density. Instead, it constrains the range of plausible densities by encoding coarse physical differences, thereby stabilizing density-oriented inference and improving interpretability. By operating at a semantic level, this representation avoids the numerical instability and hallucination issues that can arise when language models are tasked with predicting exact physical values.

\subsection{Fusion and Mass Prediction}
\label{ss:mass}
Having obtained geometry- and semantic representations aligned with the volume- and density-related factors, respectively, the remaining challenge is to integrate these heterogeneous cues into a coherent mass prediction. This integration should preserve their distinct physical roles, while remaining flexible enough to accommodate object-dependent variation in which factors dominate mass.

\subsubsection{Modality-Specific Encodings}
We encode appearance, geometry, and semantic cues using separate modality-specific networks.
The RGB image $I_o$ is encoded with \emph{DenseNet-121}~\cite{HLM17}. The point cloud $\mathbf{P}$ is encoded using \emph{PointNet}~\cite{QSM17}, preserving geometric structure relevant to volume estimation. The semantic description $t_o$ is embedded using a frozen \emph{CLIP} text encoder~\cite{RKH21}. Each encoder outputs a 512-dimensional feature vector, followed by Layer Normalization. During training, the image and geometry encoders are fine-tuned, while the text encoder remains fixed.

\subsubsection{Instance-Adaptive Fusion}
Although volume and density arise from different physical factors, their relative influence varies across objects. For example, large objects are often dominated by geometric extent, whereas for smaller objects, differences in physical composition may have a greater effect. To capture this variability, we adopt an instance-adaptive fusion strategy that modulates the contribution of each modality on a per-object basis. 

We evaluate three fusion mechanisms. The first \textbf{concatenates} modality-specific features into a single 1,536-dimensional representation, serving as a strong yet computationally efficient baseline. The second applies \textbf{self-attention} to explicitly model cross-modal interactions. Our primary design employs a \textbf{gated fusion} mechanism that predicts scalar weights for each modality and aggregates features through a weighted combination. This formulation allows the model to emphasize geometry, semantics, or appearance depending on which cues are most informative for a given object, while maintaining a clear correspondence between modalities and their physical roles.

\subsubsection{Volume and Density Regression}
From the fused representation, we predict a volume-related factor $\widehat{V}$ and a density-related factor $\hat{\rho}$ using two dedicated regression heads with identical architectures but distinct output constraints. This reflects the fact that volume and density arise from different physical factors and should be modeled separately rather than conflated in a single predictor.

The density head is implemented as a lightweight multi-layer perceptron that outputs a scalar estimate $\hat{\rho}$. We apply the custom activation function of~\cite{CM25}, which matches the empirical density distribution of the \emph{image2mass} dataset (see~\Cref{s:experiments}) and promotes physically plausible density-factor predictions, stabilizing optimization and suppressing unrealistic values, particularly in early training stages. The volume head applies a Sigmoid activation to ensure non-negative volume estimates.

The final mass prediction is obtained as $\widehat{m} = \widehat{V}\hat{\rho}$ (see Eq~\ref{eq:mass_factor}), and training is supervised solely using ground-truth mass. We employ the Absolute Log Difference Error (ALDE) loss on mass, which compares predictions on a logarithmic scale, reducing sensitivity to outliers while remaining effective across the range of object masses.

\section{Experiments}
\label{s:experiments}

\subsection{Experimental Setting}
\label{ss:setting}

\subsubsection{Datasets}
Experiments are conducted on two benchmarks for visual mass estimation. The \emph{image2mass} dataset~\cite{SSC17} is a large-scale benchmark constructed from Amazon product listings and contains 141,950 training instances and 1,435 test instances, each annotated with an RGB image, object mass, and bounding box. We also evaluate on the \emph{Amazon-Berkeley Objects (ABO)} dataset~\cite{CGD22}, which provides calibrated multi-view images and ground-truth mass for real objects. We adopt the \emph{ABO-500} subset proposed by~\cite{ZSC24}, which removes redundant instances across object classes. To ensure a fair comparison under a single-image setting, we follow the view selection protocol of~\cite{ZSC24} and use one view per object.

\subsubsection{Evaluation Metrics}
Following~\cite{SSC17}, we report Absolute Log Difference Error (ALDE), Absolute Percentage Error (APE), Minimum Ratio Error (MnRE), and the $q$-metric (off by a factor of 2). For \emph{ABO-500}, we also report Absolute Difference Error (ADE). See the supplemental material for definitions.

\subsubsection{Baselines}
We compare against three classes of non-contact, vision-based methods. All learned regression models (RGB, RGB+Depth, and ours) are trained for 25 epochs using Adam with learning rate $1\times10^{-4}$. VLM-based methods are evaluated in a zero-shot prompting setting.

\textbf{RGB-based} regressors estimate mass directly from a single RGB image. Our \emph{RGB baseline} employs a frozen \emph{ResNet-50} backbone pretrained on ImageNet, followed by three trainable fully connected layers. A Softplus activation is applied to enforce positive-valued predictions. The \emph{image2mass} algorithm~\cite{SSC17} predicts a thickness mask along with 14 auxiliary geometric features extracted from a single RGB image using an \emph{Xception} backbone~\cite{Cho17}, from which object density and volume are inferred to compute the final mass.

\textbf{RGB+Depth} methods incorporate explicit geometric cues by fusing RGB features (\emph{DenseNet}) with pseudo point clouds (\emph{PointNet}) through fully connected layers, following~\cite{CM25}.

\textbf{VLM-based} models include two open-source VLMs, \emph{Qwen2.5-VL}~\cite{YYZ24} and \emph{LLaVA}~\cite{LLW23},\footnote{\url{https://huggingface.co/llava-hf/llava-v1.6-mistral-7b-hf}} evaluated under two prompting strategies: \emph{direct} prediction, which outputs mass end-to-end, and \emph{reasoning}, which first infers material properties and conditions mass prediction on them.

\subsection{Main Results}
\label{s:mainresults}

\subsubsection{Results on \emph{image2mass}}
\Cref{t:image2mass_results} summarizes the results. Several clear trends emerge. First, VLM-based approaches perform substantially worse than all learned regression models. Second, purely RGB-based methods achieve limited accuracy, while incorporating geometric information leads to consistent improvements. Third, our proposed multi-cue model achieves the best performance across all metrics. 

\paragraph{VLM-based prediction.} Both \emph{LLaVA} and \emph{Qwen-VL} perform poorly, especially under direct prompting, exhibiting extremely large APE values and low MnRE and $q$-scores. These failures often arise from degenerate predictions (e.g., near-zero\footnote{VLMs often produce degenerate outputs such as \texttt{Mass=0.0000...}} or unrealistically large masses), indicating that current VLMs are unreliable numerical estimators of physical quantities. Stepwise reasoning improves stability but remains far inferior to learned regressors.

\paragraph{Monocular RGB vs. geometry-aware models.}
The RGB baseline achieves limited accuracy, reflecting the difficulty of inferring mass from appearance representations alone. The \emph{image2mass} model offers only modest gains despite architectural complexity. In contrast, incorporating depth-derived geometry~\cite{CM25} improves ALDE, MnRE, and $q$, highlighting the importance of geometric volume estimation.

\paragraph{Our multi-cue model.} Our framework, which jointly models geometry and material semantics, consistently outperforms all baselines. Even simple concatenation surpasses prior work, indicating that the gains stem from cue complementarity rather than fusion complexity. Gated fusion yields the best results. The learned modality weights (image: 0.13, geometry: 0.49, text: 0.36) further confirm that geometry and material dominate mass estimation.

\setlength{\tabcolsep}{0.8pt}
\begin{table}[t]
\centering
\begin{tabular}{lc S[table-format=2.3, detect-weight, mode=text] cc}
\toprule
Method
& \multicolumn{1}{c}{ALDE ($\downarrow$)}
& \multicolumn{1}{c}{APE ($\downarrow$)}
& \multicolumn{1}{c}{MnRE ($\uparrow$)}
& \multicolumn{1}{c}{Q ($\uparrow$)} \\
\midrule
\multicolumn{5}{l}{\textbf{\textit{VLM-based.}}} \\
\addlinespace[1pt]
\emph{LLaVA} (\textit{direct}) & 1.944 & 63.167 & 0.302 & 0.245 \\
\emph{LLaVA} (\textit{reasoning}) & 1.833 & 54.008 & 0.361 & 0.325 \\
\emph{Qwen-VL} (\textit{direct}) & 2.133 & 46.618 & 0.402 & 0.398 \\
\emph{Qwen-VL} (\textit{reasoning}) & 1.074 & 39.550 & 0.496 & 0.478 \\
\midrule
\multicolumn{5}{l}{\textbf{\textit{RGB-based.}}} \\
\addlinespace[1pt]
\emph{RGB baseline} & 0.843 & 1.459 & 0.517 & 0.514 \\
\emph{image2mass} & 0.655 & 1.066 & 0.579 & 0.614 \\
\midrule
\multicolumn{5}{l}{\textbf{\textit{RGB+Depth.}}} \\
\addlinespace[1pt]
Cardoso and Moreno & 0.567 & 4.886 & 0.615 & 0.686 \\
\midrule
Ours (\textit{concat.}) & \underline{0.528} & \phantom{0}\underline{\tablenum[table-format=1.3]{0.681}} & \underline{0.639} & \underline{0.717} \\
Ours (\textit{self-attn.}) & 0.560 & 0.722 & 0.627 & 0.694 \\
Ours (\textit{gated fusion}) & \textbf{0.519} & \bfseries 0.665 & \textbf{0.647} & \textbf{0.722} \\
\bottomrule
\end{tabular}
\caption{Comparison of mass prediction performance on the \emph{image2mass} test set. \textbf{Bold}: best; \underline{underlined}: second-best performance.
}
\label{t:image2mass_results}
\end{table}

\subsubsection{Results on \emph{ABO-500}}
\label{ss:abo}
\Cref{t:abo_results} reports results. As on \emph{image2mass}, VLM-based methods perform poorly, reflecting their reliance on unstable numerical prediction rather than physically grounded inference. In contrast, prior learning-based pipelines exhibit more stable behavior by implicitly or explicitly modeling geometric and material factors.

Among all \emph{single-image} methods, our model achieves the best overall performance. We also report the multi-view \emph{NeRF2Physics} pipeline, which reconstructs full 3D geometry from 30 RGB views, as a reference upper bound. Despite operating on only one view, our method remains competitive with this multi-view system and even outperforms it in terms of ADE and APE.

\setlength{\tabcolsep}{0.7pt}
\begin{table}[t]
\centering
\begin{tabular}{l S[table-format=2.3, detect-weight, mode=text] S[table-format=1.3, detect-weight, mode=text] S[table-format=2.3, detect-weight, mode=text] S[table-format=1.3, detect-weight, mode=text]}
\toprule
Method
& \multicolumn{1}{c}{ADE ($\downarrow$)}
& \multicolumn{1}{c}{ALDE ($\downarrow$)}
& \multicolumn{1}{c}{APE ($\downarrow$)}
& \multicolumn{1}{c}{MnRE ($\uparrow$)} \\
\midrule
\multicolumn{5}{l}{\textbf{\textit{VLM-based.}}} \\
\addlinespace[1pt]
\emph{LLaVA} (\textit{direct}) & 12.083 & 3.157 & 2.312 & 0.280 \\
\emph{LLaVA} (\textit{reasoning}) & 30.588 & 1.781 & 12.283 & 0.368 \\
\emph{Qwen-VL} (\textit{direct}) & 21.871 & 2.080 & 15.191 & 0.430 \\
\emph{Qwen-VL} (\textit{reasoning}) & 76.231 & 1.336 & 4.223 & 0.436 \\
\midrule
\multicolumn{5}{l}{\textbf{\textit{RGB-based.}}} \\
\addlinespace[1pt]
RGB baseline & 15.431 & 1.609 & 14.459 & 0.362 \\
\emph{image2mass} & 12.496 & 1.792 & \phantom{0}\underline{\tablenum[table-format=1.3]{0.976}} & 0.341 \\
\midrule
\multicolumn{5}{l}{\textbf{\textit{RGB+Depth.}}} \\
\addlinespace[1pt]
Cardoso and Moreno & 14.231 & 1.128 & 2.117 & 0.436 \\
\midrule
Ours (\textit{concat.}) &  10.161 & 1.034 & 1.639 & 0.407 \\
Ours (\textit{self-attn.}) &  \bfseries 7.530 & \underline{\tablenum[table-format=1.3]{1.019}} & \bfseries 0.900 & \underline{\tablenum[table-format=1.3]{0.458}} \\
Ours (\textit{gated fusion}) & \phantom{0}\underline{\tablenum[table-format=1.3]{8.321}}  & \bfseries 0.998 & 0.980 & \bfseries 0.472 \\
\midrule
\midrule
\multicolumn{5}{l}{\textbf{\textit{Multi-view input.}}} \\
\addlinespace[1pt]
\emph{NeRF2Physics} & 8.730 & 0.771 & 1.061 & 0.552\\
\bottomrule
\end{tabular}
\caption{Performance comparison on the \emph{ABO-500} test set.}
\label{t:abo_results}
\end{table}

\setlength{\tabcolsep}{3pt}
\begin{table}[t]
\centering
\begin{tabular}{ccccc S[table-format=4]}
\toprule
Category & ALDE ($\downarrow$) & APE ($\downarrow$) & MnRE ($\uparrow$) & Q ($\uparrow$) & Count\\
\midrule
Seen & 0.485 & 0.543  & 0.656 & 0.747 & 355\\
Unseen & 0.531 & 0.708  & 0.634 & 0.712 & 1080 \\
\midrule
Total & 0.519 & 0.665 & 0.647 & 0.722 & 1435 \\
\bottomrule
\end{tabular}
\caption{Performance on seen and unseen test categories.}
\label{t:seen_unseen}
\end{table}

\subsection{Generalization and Robustness}
To assess whether the observed performance gains reflect transferable physical reasoning rather than overfitting to the training distribution, we analyze category overlap between the training and test sets. Only 24.7\% of the test categories are seen during training, while the remaining 75.3\% correspond to entirely novel object categories. Moreover, even within seen categories, test instances differ substantially in appearance, geometry, color, and scale, making generalization nontrivial (see the supplemental material for examples).

Despite this limited overlap and high intra-category variability, our model exhibits strong robustness. As shown in \Cref{t:seen_unseen}, performance on unseen categories degrades only moderately relative to seen ones and follows similar trends across all metrics, suggesting that the model is not simply memorizing category-specific statistics but instead learning representations that generalize across object types.

\subsection{A Study of Material-Guided Density Factors}
A central challenge in visual mass estimation lies in modeling the influence of material-dependent density from visual input. Prior work often relies on VLMs to directly predict numerical density values, but such predictions can be unstable and physically inconsistent, especially when geometry is incomplete in a single-view setting. In contrast, our approach uses VLMs only to infer coarse material semantics (e.g., metal, wood, plastic), which are then interpreted by a learning-based model rather than treated as density values themselves. These semantic cues serve as soft physical priors that guide density estimation while avoiding brittle numeric prediction.

To better understand this design choice, we compare against two alternative density representations. First, we evaluate direct numerical density prediction by adapting \emph{NeRF2Physics} to our single-view setting (denoted \emph{*NeRF2Physics} in \Cref{t:alternative}), where a reconstructed point cloud is provided in place of its original multi-view NeRF representation. This isolates the effect of VLM-based numeric density prediction under the same geometric constraints as our method.

Unlike the original \emph{NeRF2Physics} pipeline, which relies on full 3D reconstruction from multiple views, our setting uses a point cloud reconstructed from a single RGB image. As a result, point-level features can be sparse and unreliable due to incomplete geometry and limited observations. To mitigate this issue, we also evaluate \emph{ULIP-2}~\cite{XYZ24}, a multimodal encoder jointly trained on images, text, and point clouds, which provides richer and more robust geometric--semantic representations.

Second, we consider a rule-based density baseline derived from standard physical tables. Predicted material categories are mapped to canonical density ranges from the literature, and their mean values are used as deterministic density proxies. This enforces physically plausible values but removes the ability to adapt density to object-specific context.

Results are reported in \Cref{t:alternative}. Direct VLM-based density prediction via \emph{*NeRF2Physics} performs worst, highlighting the difficulty of obtaining reliable numeric density estimates under incomplete single-view geometry. The rule-based approach improves stability but remains limited by coarse category-level averaging. In contrast, our method achieves the best performance by using material semantics as learnable priors and allowing the model to infer how they map to density-related factors in context.

Both VLM-based and rule-based methods also struggle with materials that exhibit high intra-class variability (e.g., plastic, wood, and fabric), where a single scalar cannot capture the wide range of possible physical densities. While representing density as a range can partially reflect this uncertainty, downstream mass estimation ultimately requires collapsing the range into a single value, and such averages can deviate substantially from true object densities. These findings further motivate our use of coarse semantic material descriptions combined with learning-based density inference, rather than direct numeric density prediction.

\begin{table}
    \centering
    \setlength{\tabcolsep}{4pt}
    \begin{tabular}{l cccc}
    \toprule
    Method & ALDE ($\downarrow$) & APE ($\downarrow$) & MnRE ($\uparrow$) & Q ($\uparrow$) \\
    \midrule
    \multicolumn{5}{l}{\textbf{\textit{CLIP encoder}}} \\
    \addlinespace[1pt]
    *\emph{NeRF2Physics} & 1.319 & 6.686 & 0.387 & 0.255 \\
    Rule-based & 1.301 & 6.180 & 0.390 & 0.249 \\
    \midrule
    \multicolumn{5}{l}{\textbf{\textit{ULIP encoder}}} \\
    \addlinespace[1pt]
    \emph{*NeRF2Physics} & 1.323 & 6.424 & 0.386 & 0.324 \\
    Rule-based & 1.294 & 6.165 & 0.392 & 0.333 \\
    \midrule
    Ours & 0.519 & 0.665 & 0.647 & 0.722 \\
    \bottomrule
    \end{tabular}
    \captionof{table}{Performance comparison using alternative density representations. * denotes variants operating on a 3D point cloud reconstructed from a single RGB view.}
    \label{t:alternative}
\end{table}

\begin{table}[t!]
\centering
\setlength{\tabcolsep}{0.9pt}
\begin{tabular}{ccc|cccc}
\toprule
\multicolumn{3}{c|}{Configuration} & \multirow{2}{*}{ALDE ($\downarrow$)} & \multirow{2}{*}{APE ($\downarrow$)} & \multirow{2}{*}{MnRE ($\uparrow$)} & \multirow{2}{*}{Q ($\uparrow$)}\\
\cmidrule(lr){1-3}
Image & Density & Volume & & & & \\
\midrule
\cmark & \xmark & \xmark & 0.779 & 1.199 & 0.546 & 0.570 \\
\xmark & \cmark & \xmark & 1.062 & 2.001 & 0.437 & 0.390 \\
\xmark & \xmark & \cmark & 0.641 & 0.887 & 0.587 & 0.615 \\
\midrule
\cmark & \cmark & \xmark & 0.742 & 1.245 & 0.562 & 0.589 \\
\cmark & \xmark & \cmark & 0.545 & 0.587 & 0.633 & 0.688 \\
\xmark & \cmark & \cmark & 0.591 & 0.733 & 0.608 & 0.663 \\
\midrule
\cmark & \cmark & \cmark & 0.519 & 0.665 & 0.647 & 0.722 \\
\bottomrule
\end{tabular}
\captionof{table}{{Ablation study evaluating the contribution of individual components to overall mass estimation performance.}}
\label{t:ablation}
\end{table}

\subsection{Ablation Studies}
\Cref{t:ablation} quantifies the contribution of each cue, where all inputs are ultimately derived from the same RGB image. Among single-cue variants, the geometry/volume-only model performs best, highlighting that explicit 3D structure provides the most reliable signal for mass estimation. In contrast, using material semantics alone yields the weakest performance, reflecting the inherent difficulty of inferring density-related information without geometric context.

Combining density with either appearance or geometry consistently improves results, and jointly modeling both density- and volume-related factors yields a substantial gain, confirming that the two cues provide complementary information even under a single-view setting. Adding appearance further refines the prediction, and the full model integrating Image, Density, and Volume achieves the best performance across all metrics. Overall, these results support our core design choice: explicitly separating and fusing physically distinct cues leads to more accurate and robust mass estimation.

We further test two practical variants: richer material prompts and images with unconstrained backgrounds.

\subsubsection{Multi-Material Semantic Descriptions}
Objects in everyday scenes are often composed of multiple materials. To test whether richer material cues improve density-oriented inference, we replace the single-material prompt with a multi-material prompt that asks the VLM to enumerate all visible materials, following~\cite{ZSC24}. As shown in \Cref{t:multi}, this modification does not consistently outperform the single-material variant. We attribute this to three factors: 1) an object's mass is often dominated by a single material~\cite{ZSC24}, 2) our current formulation predicts a single scalar density-related factor, which is insufficient to model mixtures of materials, and 3) longer VLM outputs may be harder to encode reliably with CLIP due to its limited context capacity; long-context text encoders such as Long-CLIP~\cite{ZZD24} may alleviate this issue. We leave explicit multi-material modeling to future work.

\setlength{\tabcolsep}{5pt}
\begin{table}[t!]
\centering
\begin{tabular}{ccccc}
\toprule
 & ALDE ($\downarrow$) & APE ($\downarrow$) & MnRE ($\uparrow$) & Q ($\uparrow$) \\
\midrule
Ours (\emph{single}) & 0.519 & 0.665 & 0.647 & 0.722 \\
Ours (\emph{multi}) & 0.532 & 0.651  & 0.640 & 0.704 \\
\bottomrule
\end{tabular}
\caption{Performance comparison between single-material and multi-material semantic descriptions on the \emph{image2mass} test set.}
\label{t:multi}
\end{table}

\subsubsection{Evaluation Beyond Segmented Benchmarks}
Both \emph{image2mass} and \emph{ABO-500} provide segmented object images with uniform backgrounds, which simplifies preprocessing and reduces background-induced ambiguity. To assess performance under less controlled visual conditions, we additionally collect six household objects and capture ten images per object under varying viewpoints. Ground-truth masses are measured manually.

\begin{figure}[t]
\centering
\setlength{\tabcolsep}{4pt}
\renewcommand{\arraystretch}{1.15}
\begin{tabular}{l@{}c@{\hspace{2pt}}c@{\hspace{2pt}}c@{\hspace{2pt}}c}
\toprule
& \includegraphics[width=0.25\linewidth]{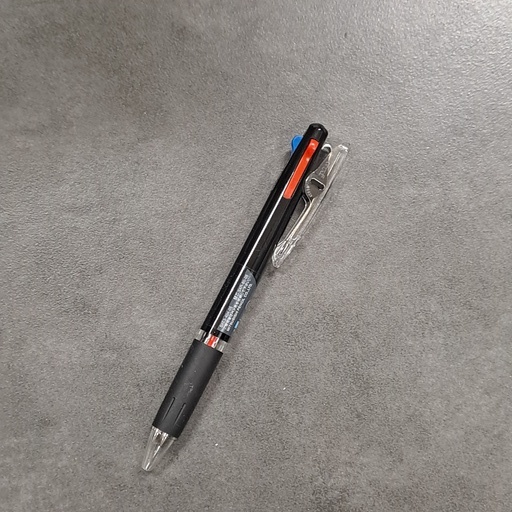}
& \includegraphics[width=0.25\linewidth]{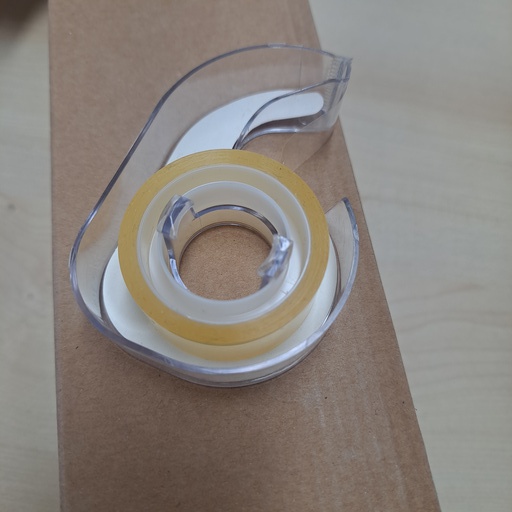}
& \includegraphics[width=0.25\linewidth]{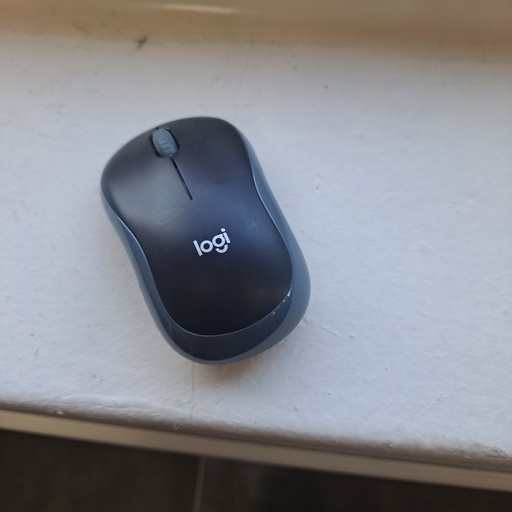} \\
\midrule
GT & \textcolor{black}{0.014} & \textcolor{black}{0.020} & \textcolor{black}{0.075} \\
Ours & 0.017 & 0.023  & 0.072 \\
\emph{image2mass} & 0.065 & 0.059  & 0.105 \\
RGB+Depth & 0.056 & 0.036 & 0.031 \\
\bottomrule
\end{tabular}
\caption{Qualitative comparison of our method with \emph{image2mass}~\protect\cite{SSC17} and the RGB+Depth approach~\protect\cite{CM25} on household-object images. Values are masses (kg).}
\label{f:real}
\end{figure}

Since our implementation (as well as the compared baselines) assumes object-centric inputs, we preprocess each image using zero-shot text-driven segmentation to isolate the target object. Specifically, we use \emph{Lang-SAM}, which integrates \emph{GroundingDINO}~\cite{LZR25} and \emph{Segment Anything}~\cite{KMR23} (see the supplemental material for details). Representative examples are shown in \Cref{f:real}. Across diverse backgrounds, our approach yields the most accurate mass estimates among single-view baselines. Full results are provided in the supplemental material.

\section{Conclusions}
In this work, we present a physically structured framework for single-image visual mass estimation grounded in the relation $m = V \cdot \rho$, using separate geometry- and material-related latent factors. Instead of relying on implicit appearance correlations or brittle end-to-end regression, we guide inference using modality-specific cues: monocular depth for object-centric geometry (volume) and VLM-derived material semantics as a density-related prior. By operating on semantic embeddings rather than raw numerical outputs, our approach leverages VLM commonsense knowledge while mitigating the brittleness of direct numeric prediction.

Our instance-adaptive fusion mechanism further improves robustness by dynamically weighting geometry, appearance, and material cues in an object-dependent manner. Experiments on \emph{image2mass} and \emph{ABO-500} demonstrate that our approach achieves state-of-the-art accuracy among single-view methods and remains competitive with multi-view pipelines that rely on full 3D reconstruction.

\clearpage
\section*{Acknowledgements}
This work was supported by the Institute of Information \& Communications Technology Planning \& Evaluation (IITP) grants (No.~RS-2022-II220290, Visual Intelligence for Space-Time Understanding and Generation, and No.~RS-2019-II191906, Artificial Intelligence Graduate School Program (POSTECH)) and by the National Research Foundation of Korea (NRF) grant (No.~RS-2024-00337559), all funded by the Korean government (MSIT).
\bibliographystyle{named}
\bibliography{MainPaper}

\clearpage
\appendix

\onecolumn
\vspace*{0.5in}

\begin{center}
{\LARGE\bf
Physically Guided Visual Mass Estimation from a Single RGB Image\\
(Supplementary Material)
}

\end{center}
\vspace{0.3in}

\section{Evaluation Metrics}
\label{s:metrics}
Let $m$ denote the ground-truth mass and $\widehat{m}$ the predicted mass. We evaluate performance using the following metrics:
\begin{itemize}
    \item \textbf{Absolute Log Difference Error (ALDE)} measures the absolute error in log space: $\text{ALDE} = \left\| \ln(m) - \ln(\widehat{m}) \right\|$.
    \item \textbf{Absolute Percentage Error (APE)} measures the relative prediction error: $\text{APE} = \cfrac{\left\| m - \widehat{m} \right\|}{m}$.
    \item \textbf{Minimum Ratio Error (MnRE)} quantifies the agreement between prediction and ground truth via their ratio, where 1 indicates a perfect prediction: $\text{MnRE} = \min \left( \cfrac{\widehat{m}}{m}, \cfrac{m}{\widehat{m}}\right)$.
    \item \textbf{q-metric} measures the agreement between prediction and ground truth by evaluating whether their ratio remains within a factor of 2: $q = \max \left( \cfrac{\widehat{m}}{m}, \cfrac{m}{\widehat{m}} \right) < 2.0$.
    \item \textbf{Absolute Difference Error (ADE)} measures the absolute error in the original mass scale: $\text{ADE} = \left\| m - \widehat{m} \right\|$.
\end{itemize}

\section{Overlapping Categories between Training and Test}
\label{s:overlap}

\begin{figure}[hbt!]
    \centering
        \centering
        \includegraphics[width=0.65\linewidth]{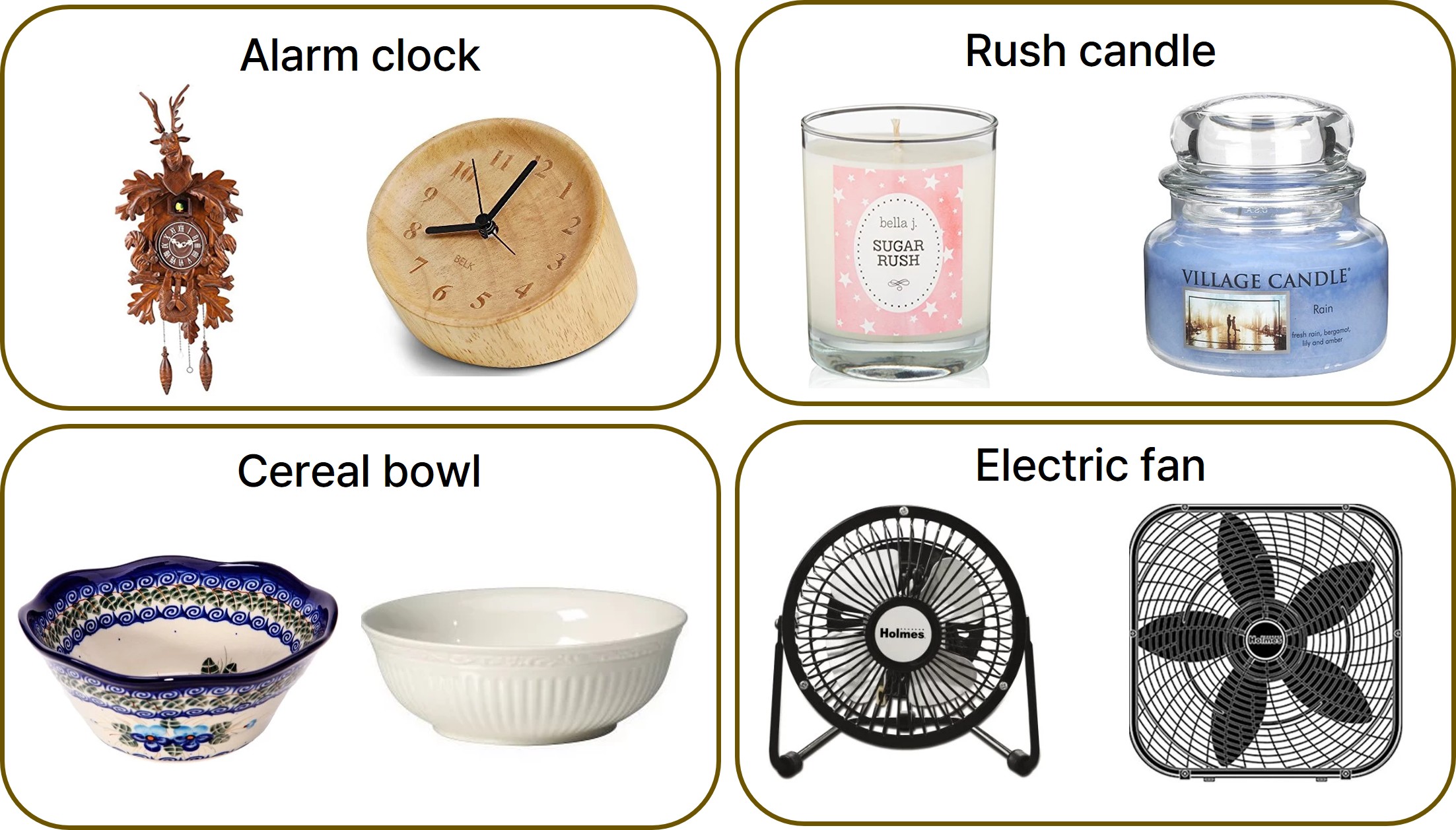}
        \captionof{figure}{Examples of overlapping categories with visual differences between training (left) and test (right).}
        \label{f:overlap}
\end{figure}

\Cref{f:overlap} shows representative examples of categories that overlap between the training and test splits of \emph{image2mass}. Although the class labels are shared, test instances often differ in appearance, geometry, color, and scale, posing nontrivial challenges for robust generalization.

\section{Qualitative Examples}
\begin{figure}[hbt!]
\centering
\setlength{\tabcolsep}{4pt}
\renewcommand{\arraystretch}{1.15}
\begin{tabular}{
    l
    @{\hspace{2pt}}S[table-format=1.3, detect-weight, mode=text ]
    @{\hspace{2pt}}S[table-format=2.3, detect-weight, mode=text]
    @{\hspace{2pt}}S[table-format=2.3, detect-weight, mode=text]
    @{\hspace{2pt}}S[table-format=2.3, detect-weight, mode=text]
    @{\hspace{2pt}}S[table-format=2.3, detect-weight, mode=text]
}
\toprule
&
{\includegraphics[width=0.154\columnwidth]{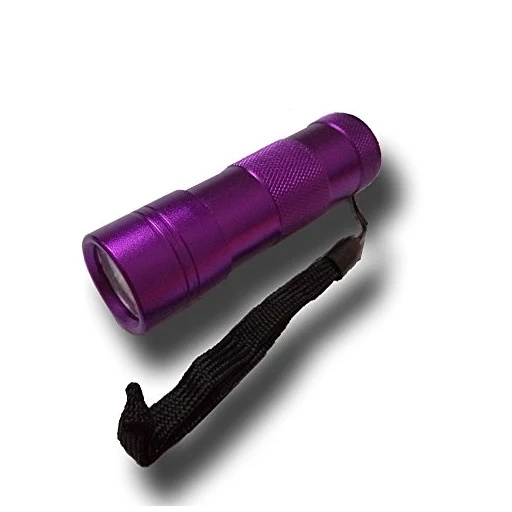}} &
{\includegraphics[width=0.154\columnwidth]{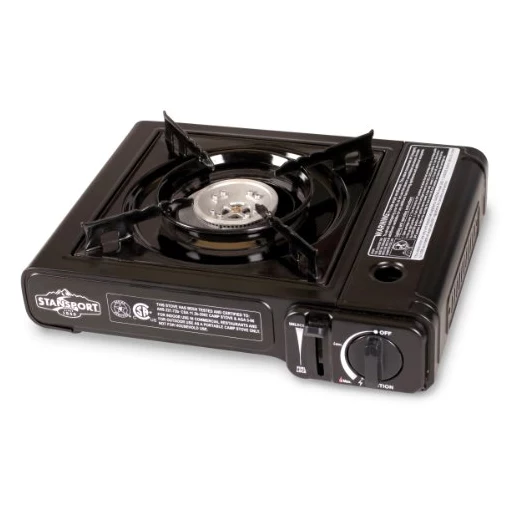}} &
{\includegraphics[width=0.154\columnwidth]{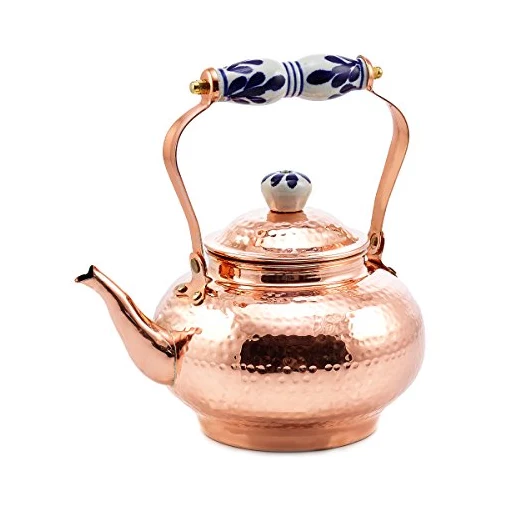}} &
{\includegraphics[width=0.154\columnwidth]{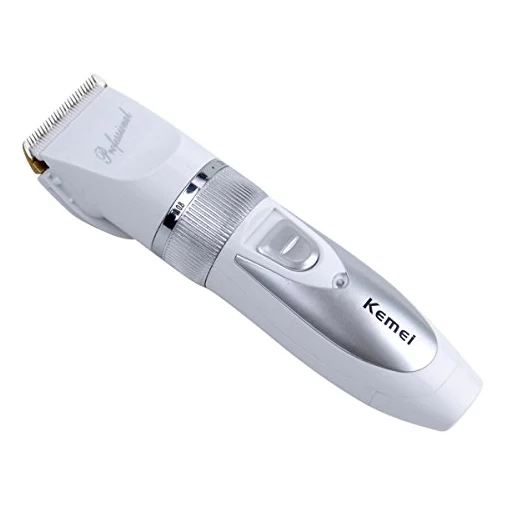}} &
{\includegraphics[width=0.154\columnwidth]{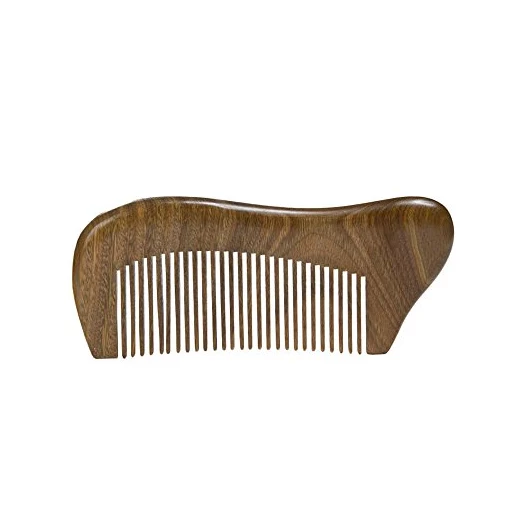}} \\
\midrule
GT & 0.063 & 1.814 & 0.907 & 0.136 & 0.059 \\
\midrule
Ours & \bfseries 0.064 & \bfseries 1.816 & \bfseries 0.905 & \phantom{0}\underline{\tablenum[table-format=1.3]{0.133}} & \phantom{0}\underline{\tablenum[table-format=1.3]{0.061}} \\
\emph{image2mass} &
0.093 & 2.558 & \phantom{0}\underline{\tablenum[table-format=1.3]{0.814}} & 0.177 & 0.037 \\
RGB+Depth & 0.082 & \phantom{0}\underline{\tablenum[table-format=1.3]{1.878}} & 0.649 & \bfseries 0.135 & \bfseries 0.060 \\
\midrule
\emph{LLaVA (direct)} & 0.001 & 12.300 & 12.300 & 0.001 & 0.001 \\
\emph{LLaVA (reasoning)} & 0.100 & 1.500 & 8.960 & 1.000 & 15.000 \\
\emph{Qwen (direct)} & \underline{0.075} & 1.500 & 1.500 & 0.100 & 0.050 \\
\emph{Qwen (reasoning)} & \underline{0.075} & 1.500 & 1.500 & 0.200 & 0.100 \\
\midrule
&
{\includegraphics[width=0.154\columnwidth]{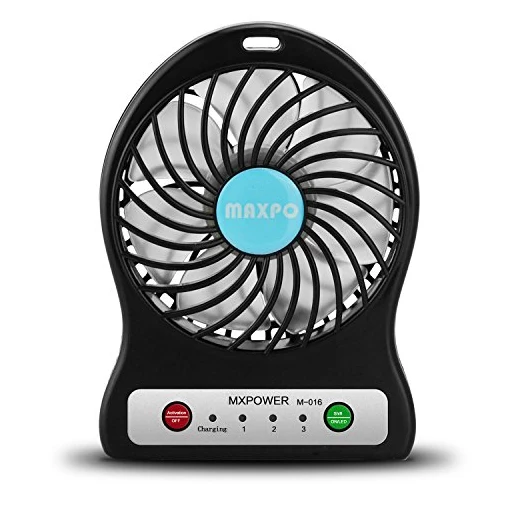}} &
{\includegraphics[width=0.154\columnwidth]{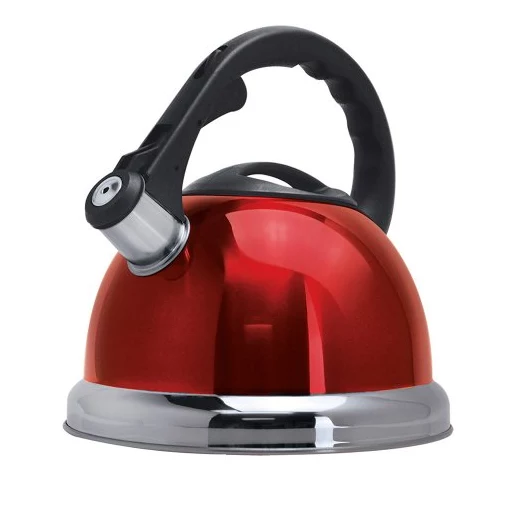}} &
{\includegraphics[width=0.154\columnwidth]{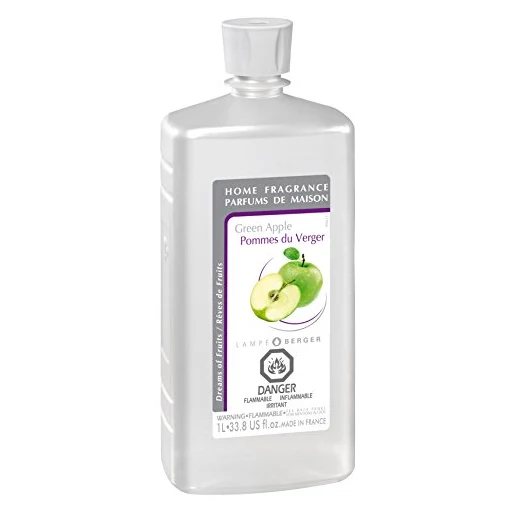}} &
{\includegraphics[width=0.154\columnwidth]{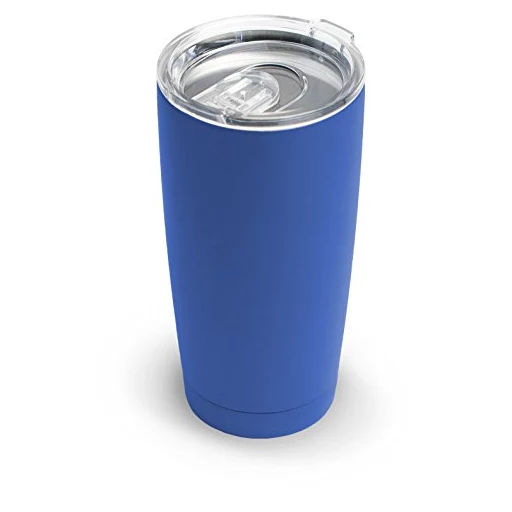}} &
{\includegraphics[width=0.154\columnwidth]{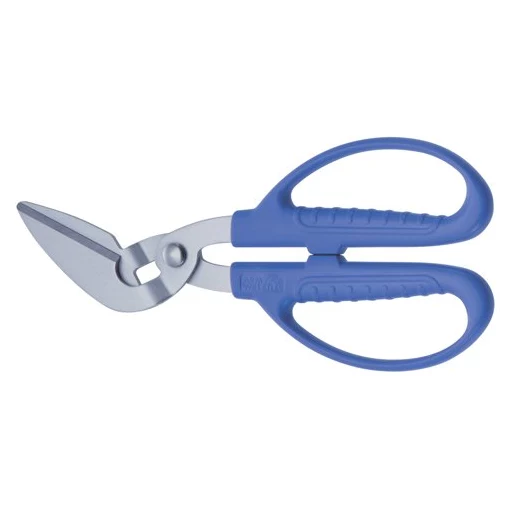}} \\
\midrule
GT & 0.149 & 0.920 & 0.907 & 0.362 & 0.131 \\
\midrule
Ours & \bfseries 0.146 & \bfseries 0.917 & \bfseries 0.912 & \phantom{0}\underline{\tablenum[table-format=1.3]{0.357}} & \bfseries 0.126 \\
\emph{image2mass} & 0.161 & \phantom{0}\underline{\tablenum[table-format=1.3]{0.937}} & \phantom{0}\underline{\tablenum[table-format=1.3]{0.898}} & \bfseries 0.365 & \phantom{0}\underline{\tablenum[table-format=1.3]{0.122}} \\
RGB+Depth & \underline{0.154} & 0.957 & 0.982 & 0.285 & 0.148 \\
\midrule
\emph{LLaVA (direct)} & 0.500 & 1.200 & 1.300 & 0.300 & 0.001 \\
\emph{LLaVA (reasoning)} & 0.500 & 1.000 & 1.500 & 41.900 & 0.150 \\
\emph{Qwen (direct)} & 0.500 & 1.500 & 1.000 & 0.500 & 0.150 \\
\emph{Qwen (reasoning)} & 0.800 & 1.500 & 1.200 & 3.850 & 0.500 \\
\bottomrule
\end{tabular}
\caption{
Qualitative comparison of our method with \emph{image2mass}~\protect\cite{SSC17}, an RGB+Depth baseline~\protect\cite{CM25}, \emph{LLaVA}~\protect\cite{LLW23}, and \emph{Qwen}~\protect\cite{YYZ24} on the \emph{image2mass} test set. For \emph{LLaVA} and \emph{Qwen}, we evaluate \emph{direct} and \emph{reasoning} strategies. Values are masses (kg). \textbf{Bold}: best; \underline{underlined}: second-best performance.}
\label{f:results}
\end{figure}

\Cref{f:results} shows example predictions on the \emph{image2mass} test set. Overall, our algorithm produces estimates that better align with object geometry and material cues across diverse categories.

\section{Evaluation Beyond Segmented Benchmarks}
Both \emph{image2mass} and \emph{ABO-500} used in the main experiments provide segmented object images with uniform backgrounds, which simplifies preprocessing and reduces background-induced ambiguity. To assess performance under less controlled visual conditions, we additionally collect six household objects and capture ten images per object under varying viewpoints. Ground-truth masses are measured manually.

\begin{figure}[hbt!]
   \centering
       \centering
       \includegraphics[width=0.6\linewidth]{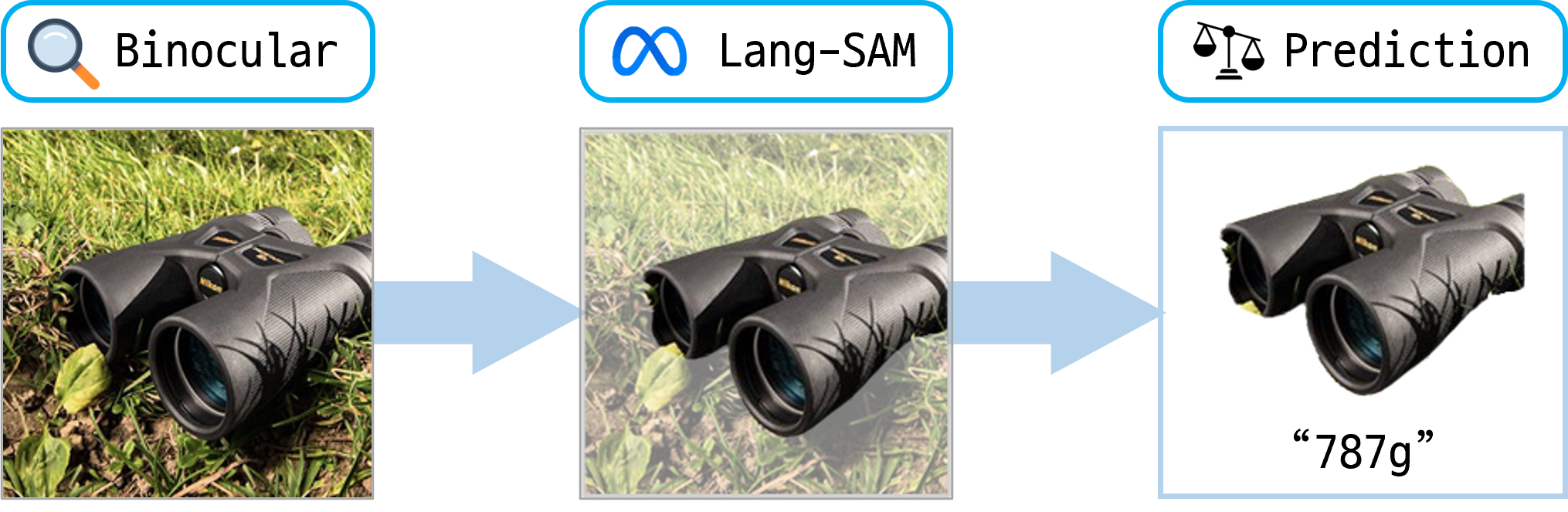}
       \captionof{figure}{
       Preprocessing workflow used for evaluation beyond segmented benchmarks. We segment the target object using a text prompt (e.g., ``Binocular'') and feed the resulting object-centric image to the mass predictor.}
       \label{f:sam}
\end{figure}

\begin{figure}[hbt!]
\centering
\setlength{\tabcolsep}{4pt}
\renewcommand{\arraystretch}{1.15}
\begin{tabular}{l@{}c@{\hspace{2pt}}c@{\hspace{2pt}}c@{\hspace{2pt}}c@{\hspace{2pt}}c}
\toprule
&
\includegraphics[width=0.154\columnwidth]{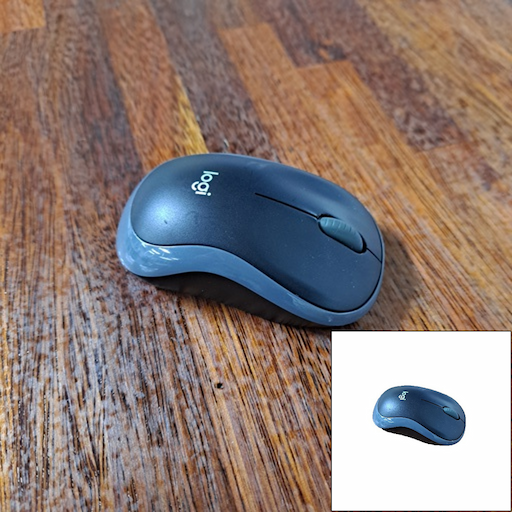} &
\includegraphics[width=0.154\columnwidth]{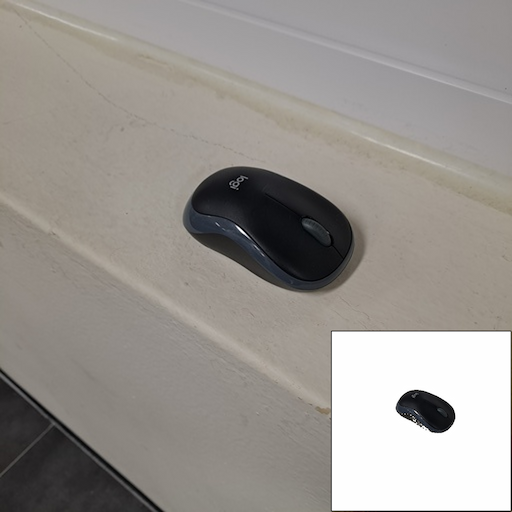} &
\includegraphics[width=0.154\columnwidth]{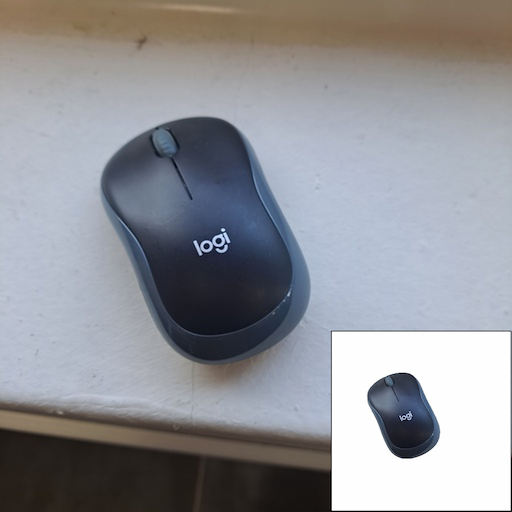} &
\includegraphics[width=0.154\columnwidth]{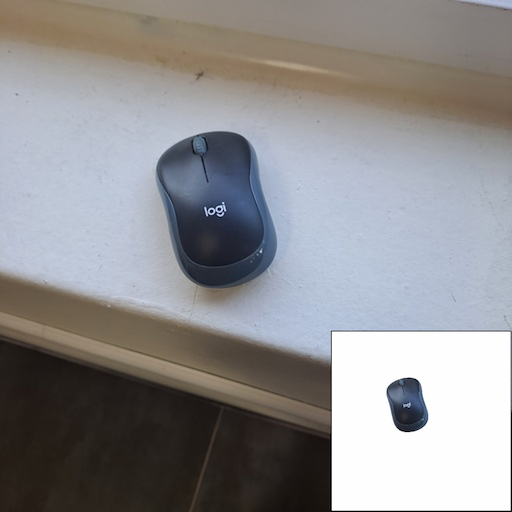} &
\includegraphics[width=0.154\columnwidth]{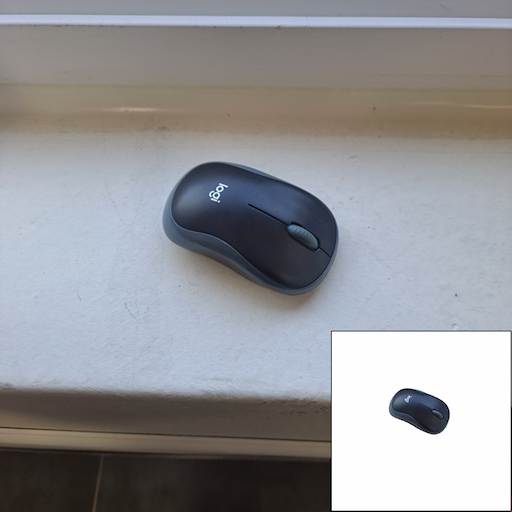} \\
\midrule
GT & 0.075 & 0.075 & 0.075 & 0.075 & 0.075 \\
\midrule
Ours & \textbf{0.066} & \textbf{0.072} & \textbf{0.072} & \textbf{0.077} & \textbf{0.074} \\
\emph{image2mass} & \underline{0.094} & 0.166 & \underline{0.105} & 0.136 & 0.113 \\
RGB+Depth & 0.029 & 0.029 & 0.031 & 0.028 & 0.025 \\
\midrule
\emph{LLaVA (direct)} & 0.001 & 0.001 & 0.001 & 0.001 & 0.001 \\
\emph{LLaVA (reasoning)} & 0.500 & \underline{0.100} & 0.500 & 0.500 & \underline{0.100} \\
\emph{Qwen (direct)} & 0.150 & \underline{0.100} & 0.150 & \underline{0.100} & 0.150 \\
\emph{Qwen (reasoning)} & 0.150 & 0.150 & 0.150 & \underline{0.100} & \underline{0.100} \\
\midrule
&
\includegraphics[width=0.154\columnwidth]{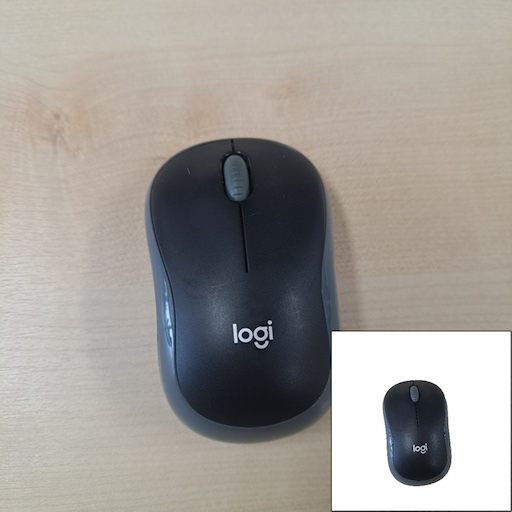} &
\includegraphics[width=0.154\columnwidth]{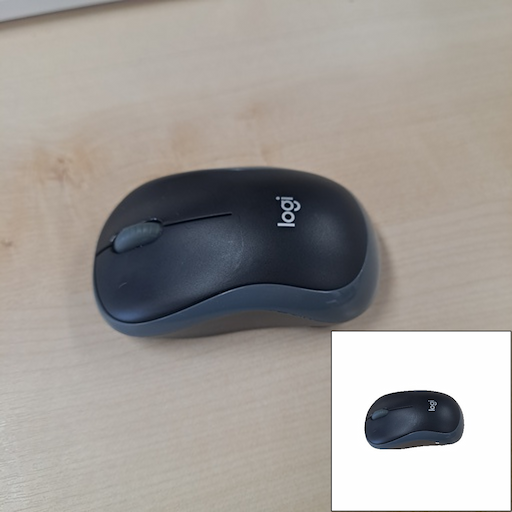} &
\includegraphics[width=0.154\columnwidth]{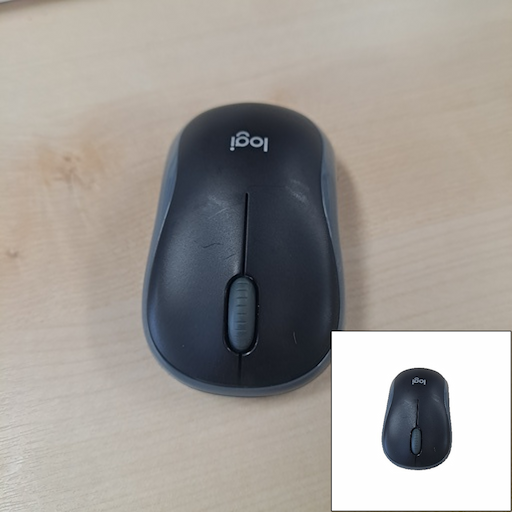} &
\includegraphics[width=0.154\columnwidth]{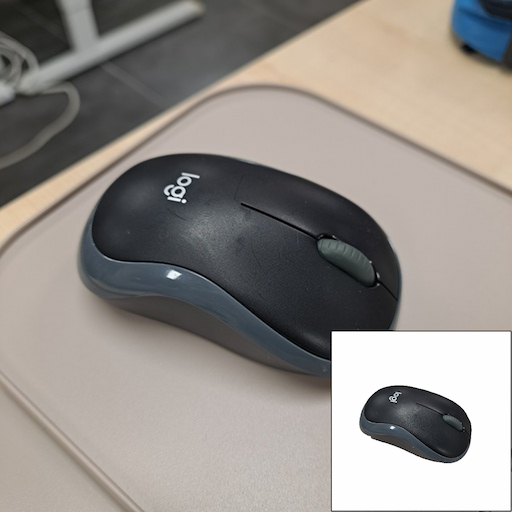} &
\includegraphics[width=0.154\columnwidth]{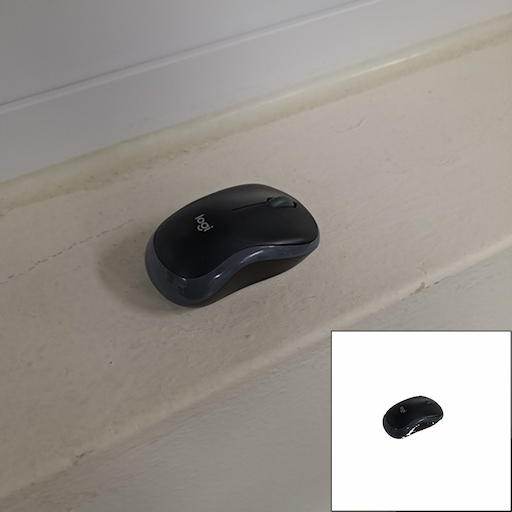} \\
\midrule
GT & 0.075 & 0.075 & 0.075 & 0.075 & 0.075 \\
\midrule
Ours & \textbf{0.072} &\textbf{0.078} & \textbf{0.075} & \textbf{0.063} & \textbf{0.065} \\
\emph{image2mass} & \underline{0.093} & \underline{0.100} & \underline{0.087} & 0.109 & 0.156 \\
RGB+Depth & 0.043 & 0.025 & 0.033 & 0.028 & 0.026 \\
\midrule
\emph{LLaVA (direct)} & 0.001 & 0.001 & 0.001 & 0.010 & 0.001 \\
\emph{LLaVA (reasoning)} & 0.500 & 0.150 & 0.500 & 0.500 & 0.150 \\
\emph{Qwen (direct)} & 0.150 & 0.150 & 0.150 & \underline{0.100} & \underline{0.100} \\
\emph{Qwen (reasoning)} & 0.150 & 0.150 & 0.150 & 0.150 & \underline{0.100} \\
\bottomrule
\end{tabular}
\caption{Qualitative comparison on household-object images. Values are masses (kg). For each example, the bottom-right shows the segmented object-centric image used for mass prediction. Rows (top to bottom): Ground truth (GT), ours, \emph{image2mass}~\protect\cite{SSC17}, RGB+Depth~\protect\cite{CM25}, \emph{LLaVA}~\protect\cite{LLW23}, and \emph{Qwen-VL}~\protect\cite{YYZ24}. For \emph{LLaVA} and \emph{Qwen}, we evaluate \emph{direct} and \emph{reasoning} strategies. \textbf{Bold}: best; \underline{underlined}: second-best performance.}
\label{f:real1}
\end{figure}

\begin{figure}[hbt!]
\centering
\setlength{\tabcolsep}{4pt}
\renewcommand{\arraystretch}{1.15}
\begin{tabular}{l@{}c@{\hspace{2pt}}c@{\hspace{2pt}}c@{\hspace{2pt}}c@{\hspace{2pt}}c}
\toprule
&
\includegraphics[width=0.154\columnwidth]{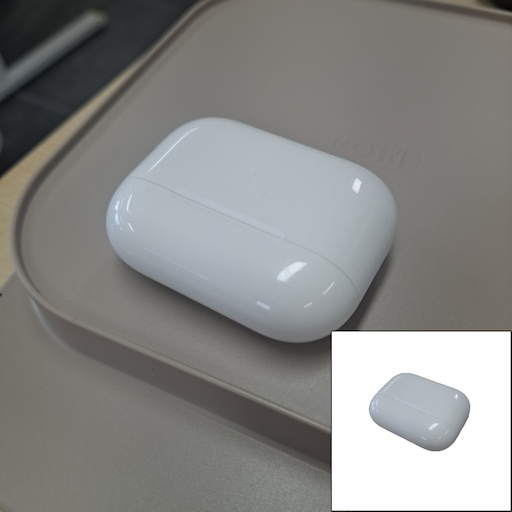} &
\includegraphics[width=0.154\columnwidth]{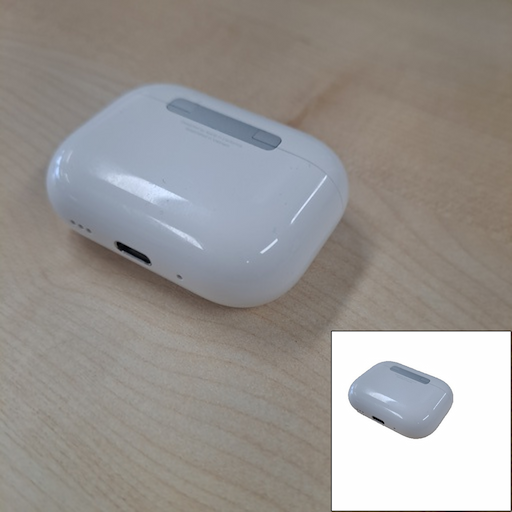} &
\includegraphics[width=0.154\columnwidth]{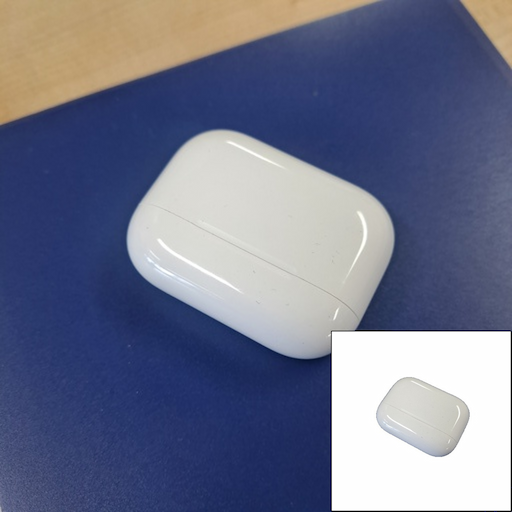} &
\includegraphics[width=0.154\columnwidth]{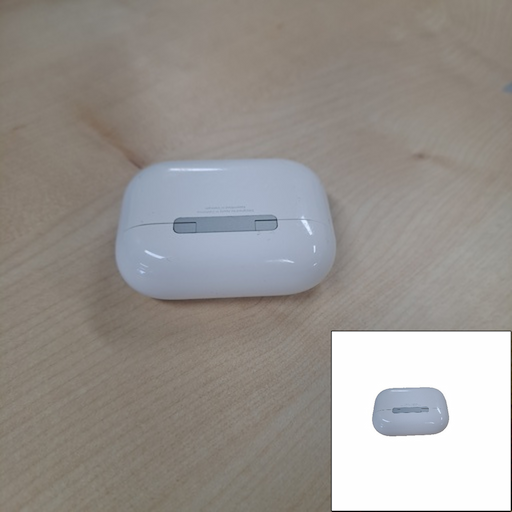} &
\includegraphics[width=0.154\columnwidth]{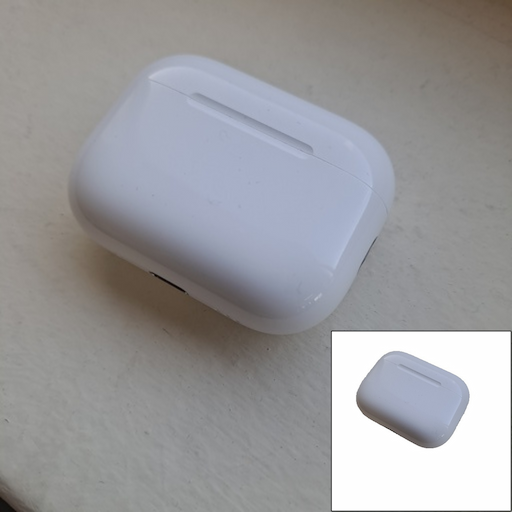} \\
\midrule
GT & 0.056 & 0.056 & 0.056 & 0.056 & 0.056 \\
\midrule
Ours & \textbf{0.061} & \textbf{0.061} & \textbf{0.059} & \textbf{0.062} & \textbf{0.049} \\
\emph{image2mass} & 0.043 & 0.041 & \underline{0.043} & \underline{0.043} & \underline{0.043} \\
RGB+Depth & 0.070 & \underline{0.067} & 0.028 & 0.040 & 0.037 \\
\midrule
\emph{LLaVA (direct)} & 0.001 & 0.001 & 0.001 & 0.001 & 0.001 \\
\emph{LLaVA (reasoning)} & 0.100 & 0.100 & 0.100 & 0.500 & 0.750 \\
\emph{Qwen (direct)} & 0.045 & 0.200 & 0.100 & 0.500 & 0.200 \\
\emph{Qwen (reasoning)} & \underline{0.050} & 0.100 & 0.150 & 0.100 & 0.252 \\
\midrule
&
\includegraphics[width=0.154\columnwidth]{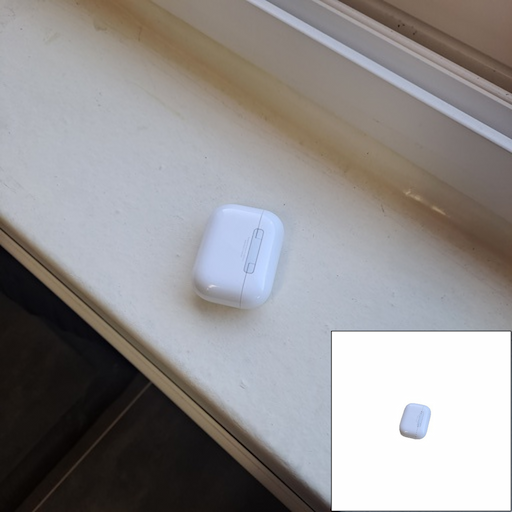} &
\includegraphics[width=0.154\columnwidth]{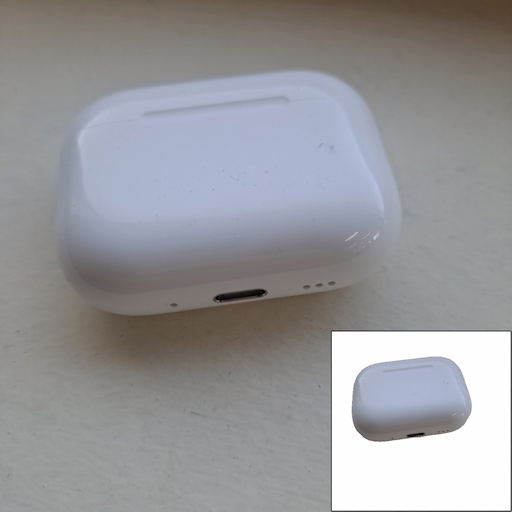} &
\includegraphics[width=0.154\columnwidth]{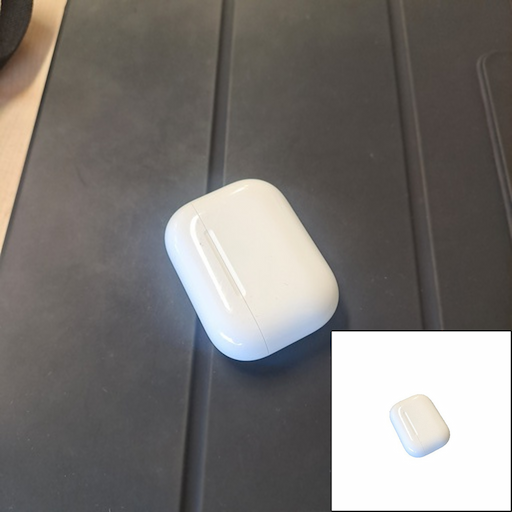} &
\includegraphics[width=0.154\columnwidth]{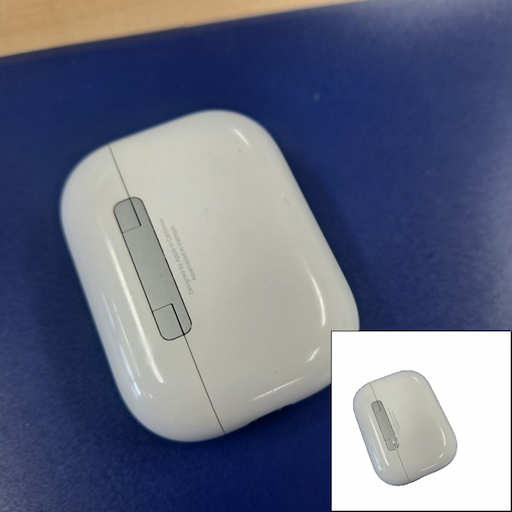} &
\includegraphics[width=0.154\columnwidth]{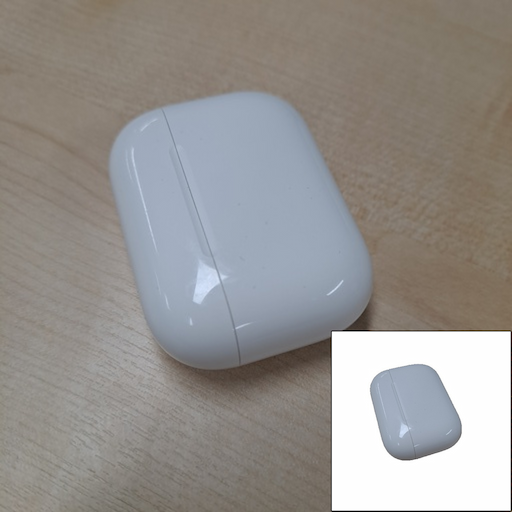} \\
\midrule
GT & 0.056 & 0.056 & 0.056 & 0.056 & 0.056 \\
\midrule
Ours & \textbf{0.063} & \textbf{0.048} & \textbf{0.061} & \underline{0.053} & \textbf{0.062} \\
\emph{image2mass} & 0.039 & 0.042 & \underline{0.043} & 0.050 & 0.043 \\
RGB+Depth & \underline{0.047} & \underline{0.068} & 0.023 & \textbf{0.054} & \underline{0.049} \\
\midrule
\emph{LLaVA (direct)} & 0.001 & 0.001 & 0.001 & 0.001 & 0.001 \\
\emph{LLaVA (reasoning)} & 0.150 & 0.150 & 0.100 & 0.500 & 0.010 \\
\emph{Qwen (direct)} & 0.100 & 0.100 & 0.025 & 0.500 & 0.045 \\
\emph{Qwen (reasoning)} & 0.100 & 0.100 & 0.150 & 0.500 & 0.126 \\
\bottomrule
\end{tabular}
\caption{Qualitative comparison on household-object images (continued).}
\label{f:real2}
\end{figure}

\begin{figure}[hbt!]
\centering
\setlength{\tabcolsep}{4pt}
\renewcommand{\arraystretch}{1.15}
\begin{tabular}{
    l
    @{\hspace{2pt}}S[table-format=2.3, detect-weight, mode=text]
    @{\hspace{2pt}}S[table-format=2.3, detect-weight, mode=text]
    @{\hspace{2pt}}S[table-format=2.3, detect-weight, mode=text]
    @{\hspace{2pt}}S[table-format=2.3, detect-weight, mode=text]
    @{\hspace{2pt}}S[table-format=1.3, detect-weight, mode=text]
}
\toprule
&
{\includegraphics[width=0.154\columnwidth]{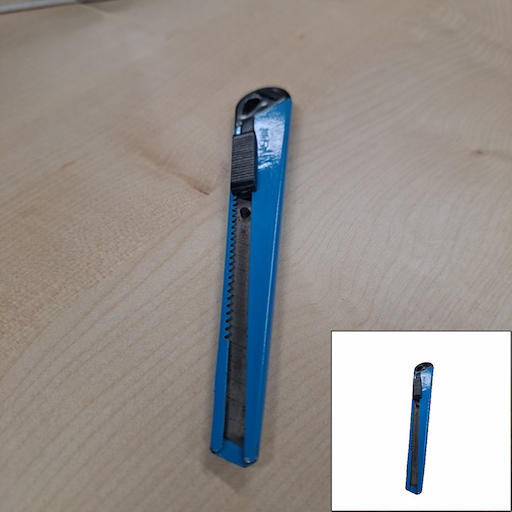}} &
{\includegraphics[width=0.154\columnwidth]{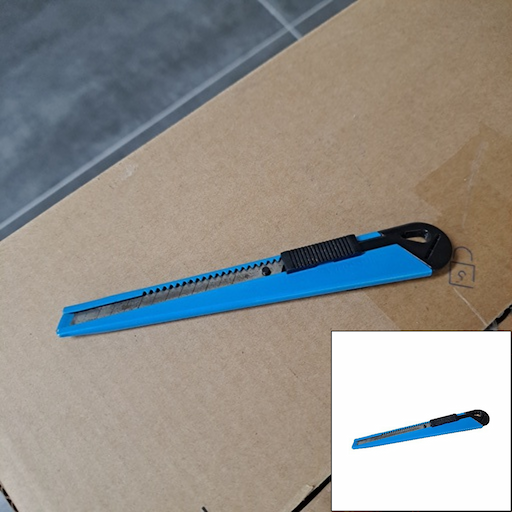}} &
{\includegraphics[width=0.154\columnwidth]{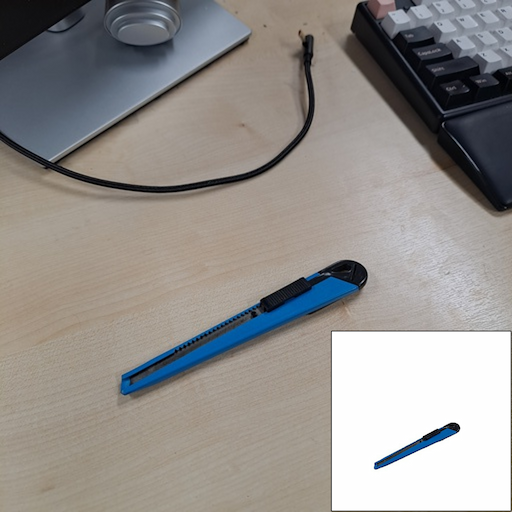}} &
{\includegraphics[width=0.154\columnwidth]{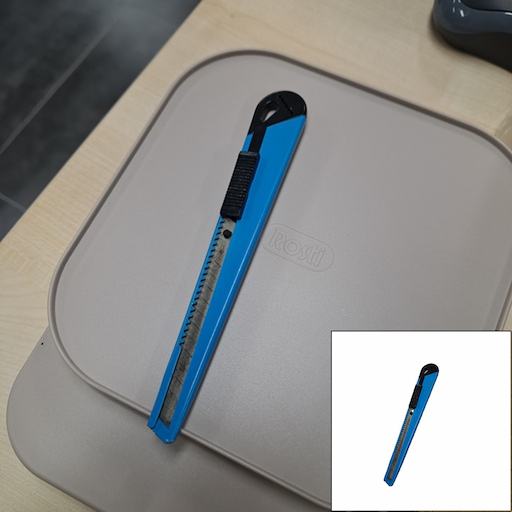}} &
{\includegraphics[width=0.154\columnwidth]{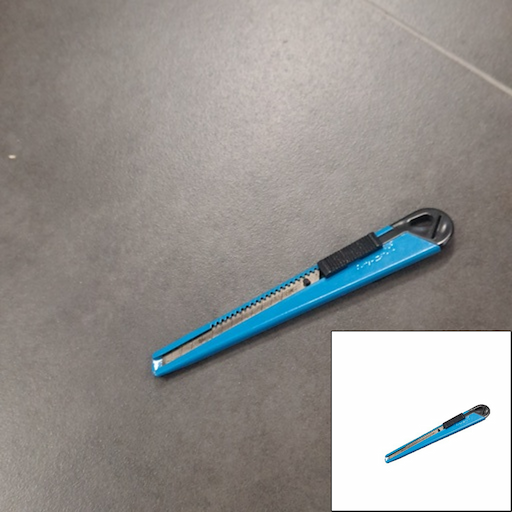}} \\
\midrule
GT & 0.024 & 0.024 & 0.024 & 0.024 & 0.024 \\
\midrule
Ours & \bfseries 0.026 & \bfseries 0.027 & \bfseries 0.023 & \bfseries 0.021 & \underline{0.022} \\
\emph{image2mass} & \phantom{0}\underline{\tablenum[table-format=1.3]{0.030}} & 0.032 & 0.050 & \phantom{0}\underline{\tablenum[table-format=1.3]{0.028}} & 0.029 \\
RGB+Depth & 0.065 & \phantom{0}\underline{\tablenum[table-format=1.3]{0.031}} & \phantom{0}\underline{\tablenum[table-format=1.3]{0.027}} & 0.066 & \bfseries 0.025 \\
\midrule
\emph{LLaVA (direct)} & 12.300 & 12.300 & 0.001 & 0.001 & 0.001 \\
\emph{LLaVA (reasoning)} & 61.670 & 0.150 & 0.010 & 12.300 & 0.010 \\
\emph{Qwen (direct)} & 0.100 & 0.150 & 0.010 & 0.050 & 0.100 \\
\emph{Qwen (reasoning)} & 0.050 & 0.050 & 0.100 & 0.050 & 0.050 \\
\midrule
&
{\includegraphics[width=0.154\columnwidth]{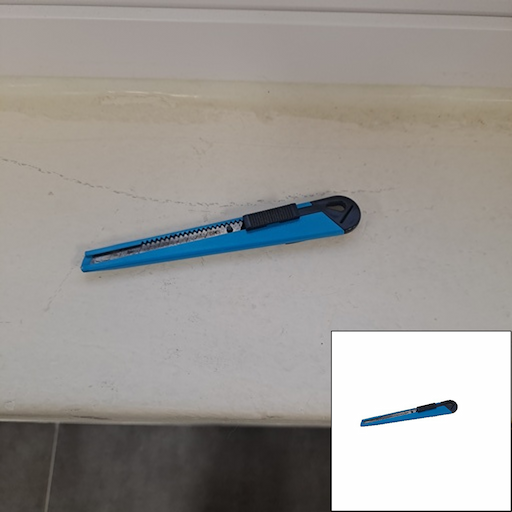}} &
{\includegraphics[width=0.154\columnwidth]{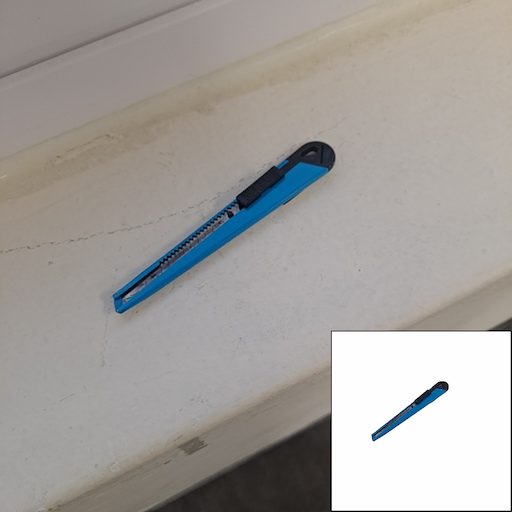}} &
{\includegraphics[width=0.154\columnwidth]{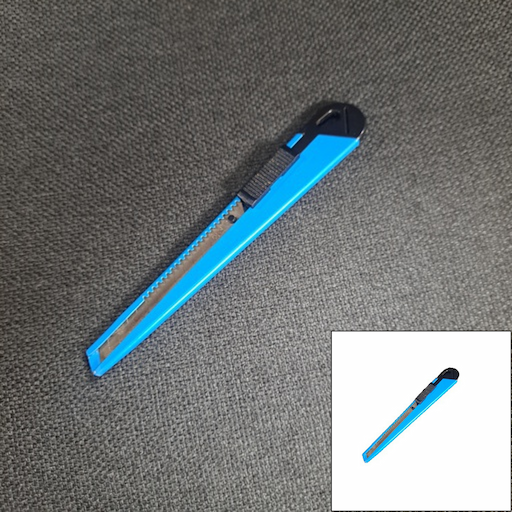}} &
{\includegraphics[width=0.154\columnwidth]{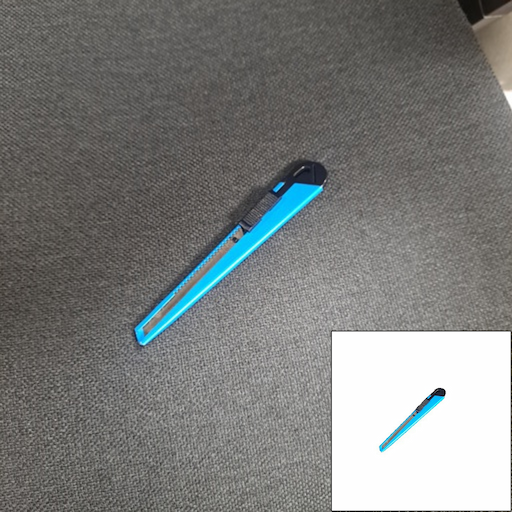}} &
{\includegraphics[width=0.154\columnwidth]{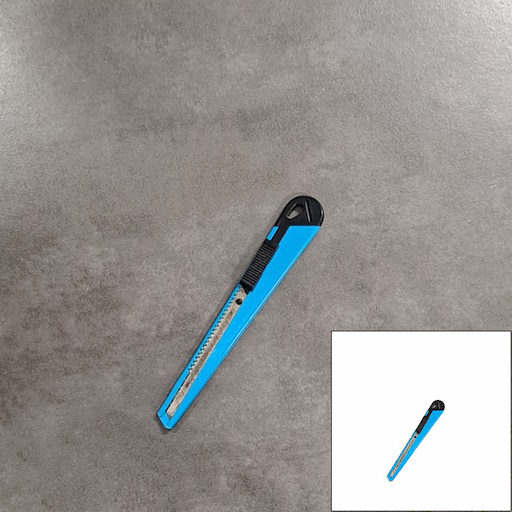}} \\
\midrule
GT & 0.024 & 0.024 & 0.024 & 0.024 & 0.024 \\
\midrule
Ours & \bfseries 0.026 & \phantom{0}\underline{\tablenum[table-format=1.3]{0.020}} & \bfseries 0.024 & \bfseries 0.023 & \bfseries 0.021 \\
\emph{image2mass} & \phantom{0}\underline{\tablenum[table-format=1.3]{0.030}} & 0.075 & \phantom{0}\underline{\tablenum[table-format=1.3]{0.022}} & 0.028 & \underline{0.030} \\
RGB+Depth & \bfseries 0.022 & \bfseries 0.025 & \phantom{0}\underline{\tablenum[table-format=1.3]{0.026}} & \phantom{0}\underline{\tablenum[table-format=1.3]{0.027}} & \bfseries 0.021 \\
\midrule
\emph{LLaVA (direct)} & 12.300 & 0.001 & 12.300 & 12.300 & 0.001 \\
\emph{LLaVA (reasoning)} & 1.000 & 0.100 & 0.785 & 1.000 & 0.100 \\
\emph{Qwen (direct)} & 0.100 & \phantom{0}\underline{\tablenum[table-format=1.3]{0.020}} & 0.100 & 0.100 & 0.100 \\
\emph{Qwen (reasoning)} & 0.075 & \phantom{0}\underline{\tablenum[table-format=1.3]{0.020}} & 0.050 & 0.050 & 0.050 \\
\bottomrule
\end{tabular}
\caption{Qualitative comparison on household-object images (continued).}
\label{f:real3}
\end{figure}

\begin{figure}[hbt!]
\centering
\setlength{\tabcolsep}{4pt}
\renewcommand{\arraystretch}{1.15}
\begin{tabular}{
    l
    @{\hspace{2pt}}S[table-format=1.3, detect-weight, mode=text]
    @{\hspace{2pt}}S[table-format=1.3, detect-weight, mode=text]
    @{\hspace{2pt}}S[table-format=1.3, detect-weight, mode=text]
    @{\hspace{2pt}}S[table-format=2.3, detect-weight, mode=text]
    @{\hspace{2pt}}S[table-format=1.3, detect-weight, mode=text]
}
\toprule
&
{\includegraphics[width=0.154\columnwidth]{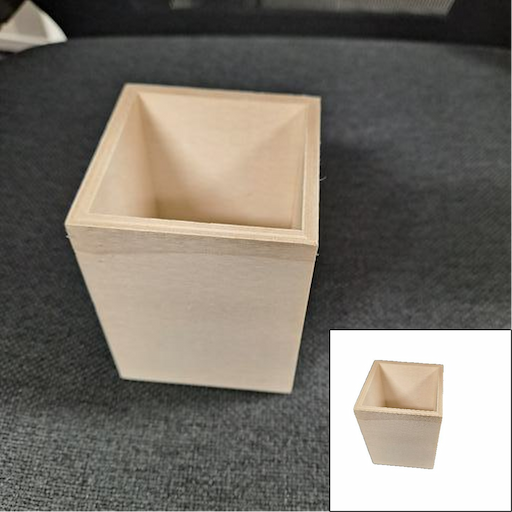}} &
{\includegraphics[width=0.154\columnwidth]{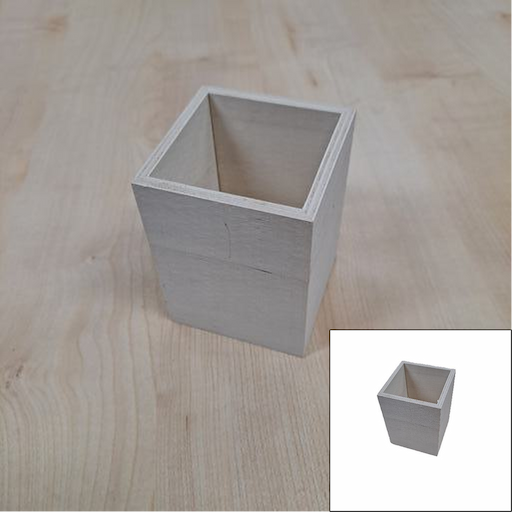}} &
{\includegraphics[width=0.154\columnwidth]{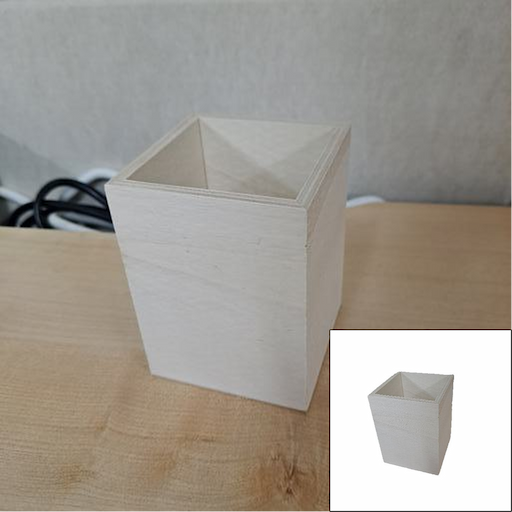}} &
{\includegraphics[width=0.154\columnwidth]{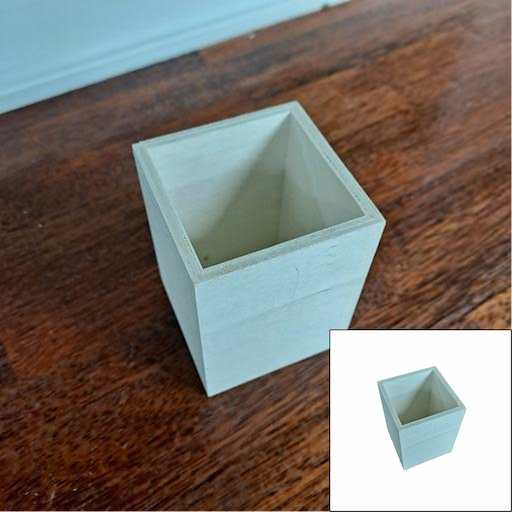}} &
{\includegraphics[width=0.154\columnwidth]{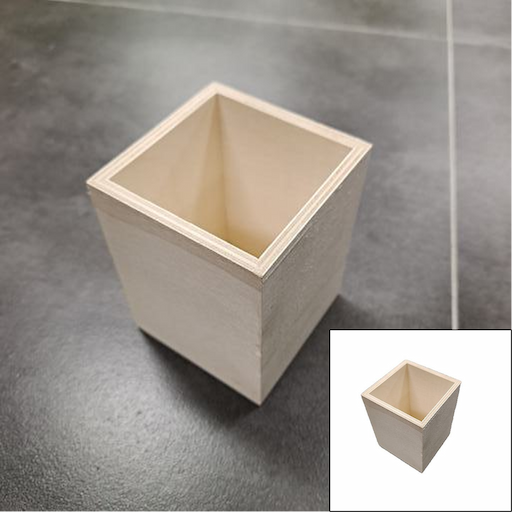}} \\
\midrule
GT & 0.081 & 0.081 & 0.081 & 0.081 & 0.081 \\
\midrule
Ours & \underline{\tablenum[table-format=1.3]{0.094}} & \bfseries 0.080 & \underline{0.087} & \bfseries 0.087 & \bfseries 0.077 \\
\emph{image2mass} & 0.194 & 0.210 & 0.217 & 0.205 & 0.157 \\
RGB+Depth & \bfseries 0.088 & \underline{0.084} & \bfseries 0.082 & \phantom{0}\underline{\tablenum[table-format=1.3]{0.099}} & \underline{0.113} \\
\midrule
\emph{LLaVA (direct)} & 0.001 & 0.001 & 0.001 & 0.001 & 0.001 \\
\emph{LLaVA (reasoning)} & 5.000 & 5.000 & 1.500 & 0.300 & 1.000 \\
\emph{Qwen (direct)} & 0.001 & 0.001 & 0.001 & 0.100 & 0.001 \\
\emph{Qwen (reasoning)} & 0.500 & 0.650 & 0.650 & 0.500 & 0.450 \\
\midrule
&
{\includegraphics[width=0.154\columnwidth]{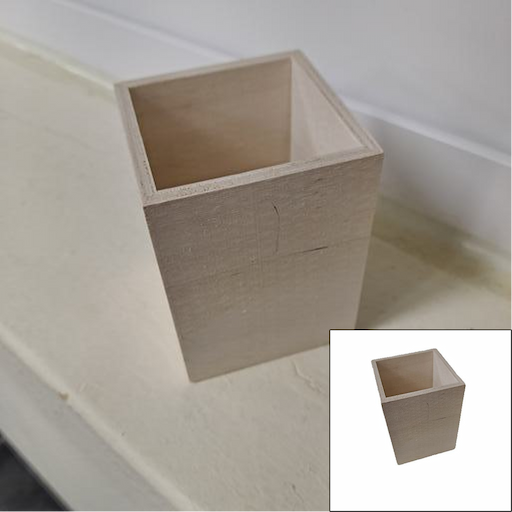}} &
{\includegraphics[width=0.154\columnwidth]{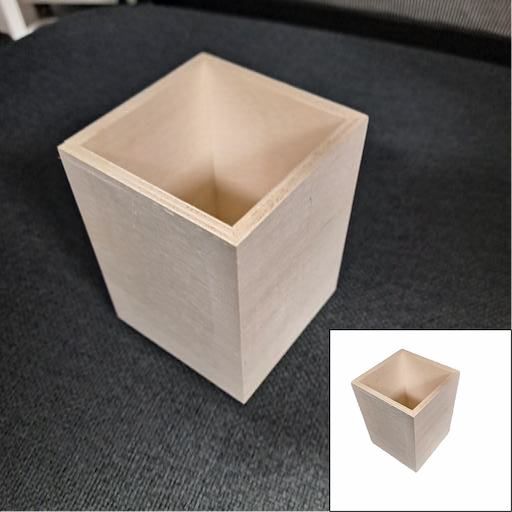}} &
{\includegraphics[width=0.154\columnwidth]{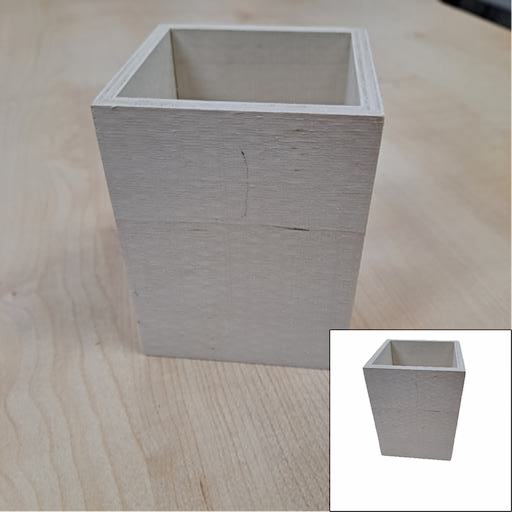}} &
{\includegraphics[width=0.154\columnwidth]{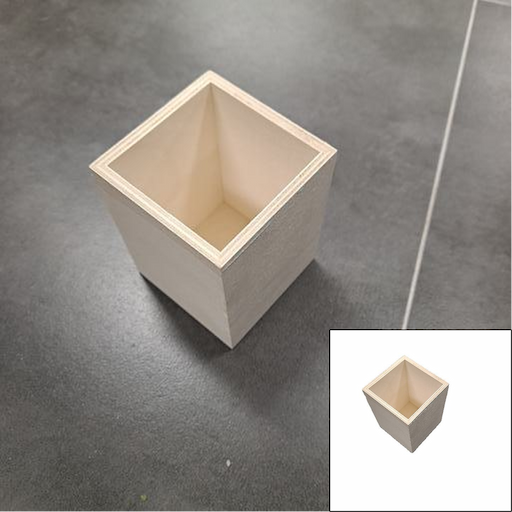}} &
{\includegraphics[width=0.154\columnwidth]{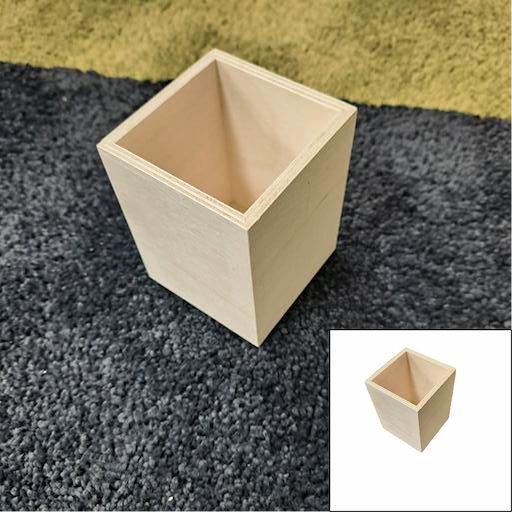}} \\
\midrule
GT & 0.081 & 0.081 & 0.081 & 0.081 & 0.081 \\
\midrule
Ours & \bfseries 0.083 & \bfseries 0.092 & \underline{0.075} & \bfseries 0.084 & \bfseries 0.088 \\
\emph{image2mass} & 0.203 & 0.189 & 0.222 & 0.194 & 0.186 \\
RGB+Depth & \underline{0.121} & \underline{0.107} & \bfseries 0.081 & \phantom{0}\underline{\tablenum[table-format=1.3]{0.057}} & \underline{0.092} \\
\midrule
\emph{LLaVA (direct)} & 0.001 & 0.001 & 0.001 & 0.001 & 0.001 \\
\emph{LLaVA (reasoning)} & 0.500 & 2.000 & 0.100 & 0.300 & 5.000 \\
\emph{Qwen (direct)} & 0.001 & 0.001 & 0.001 & 0.001 & 0.001 \\
\emph{Qwen (reasoning)} & 0.900 & 0.450 & 1.500 & 12.150 & 0.550 \\
\bottomrule
\end{tabular}
\caption{Qualitative comparison on household-object images (continued).}
\label{f:real4}
\end{figure}

\begin{figure}[hbt!]
\centering
\setlength{\tabcolsep}{4pt}
\renewcommand{\arraystretch}{1.15}
\begin{tabular}{
    l
    @{\hspace{2pt}}S[table-format=2.3, detect-weight, mode=text]
    @{\hspace{2pt}}S[table-format=1.3, detect-weight, mode=text]
    @{\hspace{2pt}}S[table-format=1.3, detect-weight, mode=text]
    @{\hspace{2pt}}S[table-format=2.3, detect-weight, mode=text]
    @{\hspace{2pt}}S[table-format=1.3, detect-weight, mode=text]
}
\toprule
&
{\includegraphics[width=0.154\columnwidth]{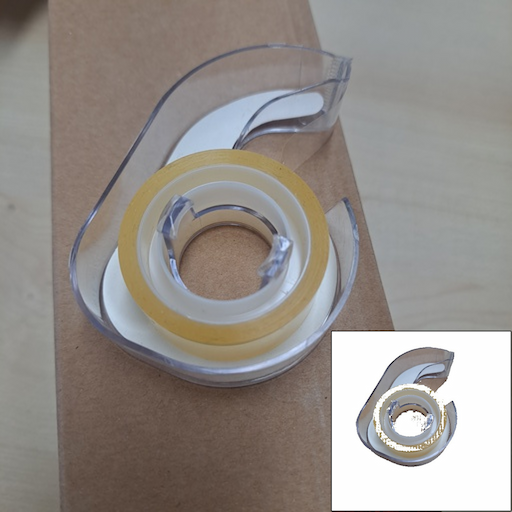}} &
{\includegraphics[width=0.154\columnwidth]{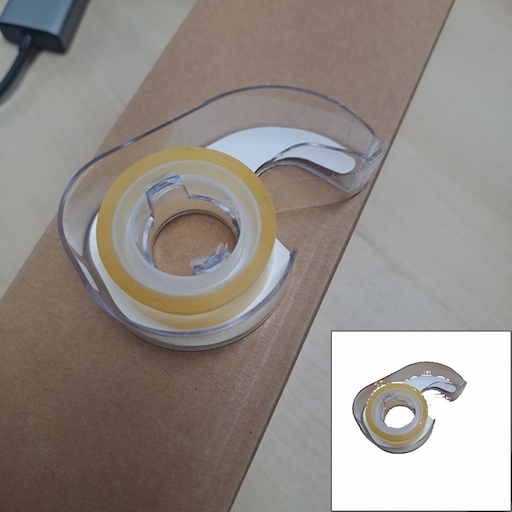}} &
{\includegraphics[width=0.154\columnwidth]{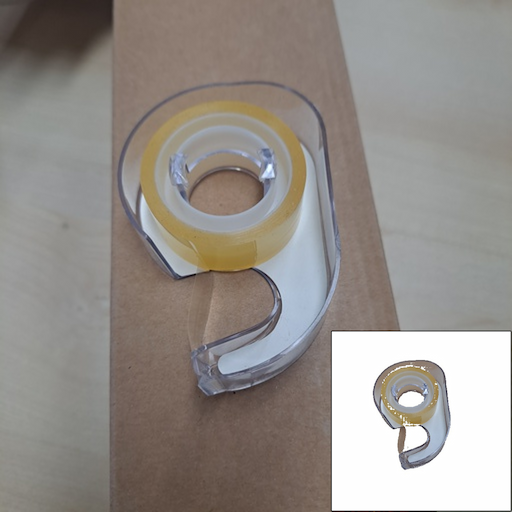}} &
{\includegraphics[width=0.154\columnwidth]{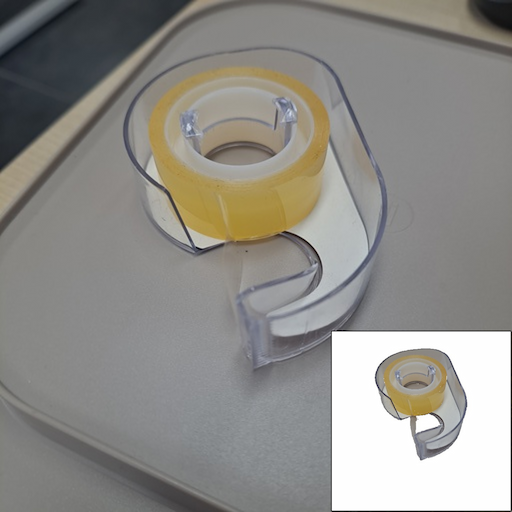}} &
{\includegraphics[width=0.154\columnwidth]{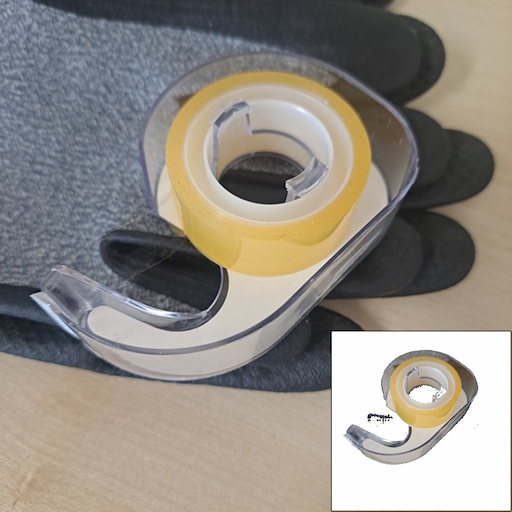}} \\
\midrule
GT & 0.020 & 0.020 & 0.020 & 0.020 & 0.020 \\
\midrule
Ours & \bfseries 0.023 & \bfseries 0.025 & \bfseries 0.023 & \phantom{0}\underline{\tablenum[table-format=1.3]{0.032}} & \underline{0.036} \\
\emph{image2mass} & 0.059 & 0.050 & 0.046 & 0.041 & 0.052 \\
RGB+Depth & \phantom{0}\underline{\tablenum[table-format=1.3]{0.036}} & \bfseries 0.025 & \underline{0.035} & 0.066 & 0.062 \\
\midrule
\emph{LLaVA (direct)} & 0.001 & 0.001 & 0.001 & 0.001 & 0.001 \\
\emph{LLaVA (reasoning)} & 0.001 & 0.001 & 0.001 & 0.079 & 0.001 \\
\emph{Qwen (direct)} & 0.500 & 0.100 & 0.100 & 0.100 & 0.100 \\
\emph{Qwen (reasoning)} & 10.000 & \underline{0.010} & 0.100 & \bfseries 0.010 & \bfseries 0.010 \\
\midrule
&
{\includegraphics[width=0.154\columnwidth]{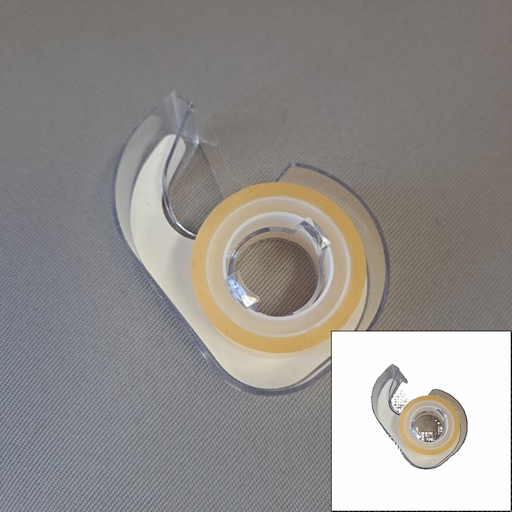}} &
{\includegraphics[width=0.154\columnwidth]{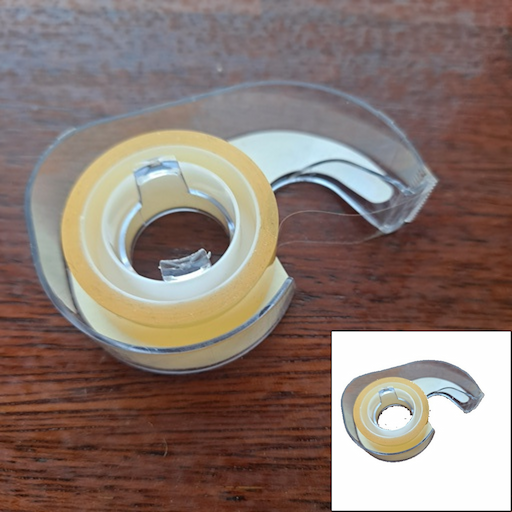}} &
{\includegraphics[width=0.154\columnwidth]{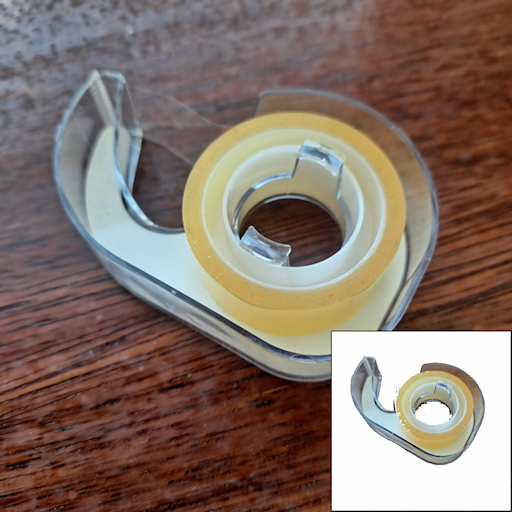}} &
{\includegraphics[width=0.154\columnwidth]{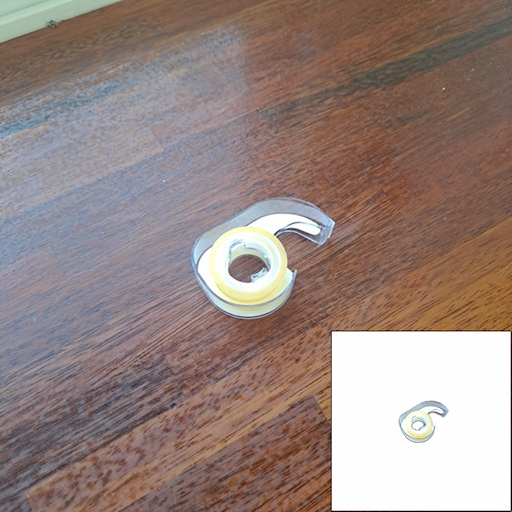}} &
{\includegraphics[width=0.154\columnwidth]{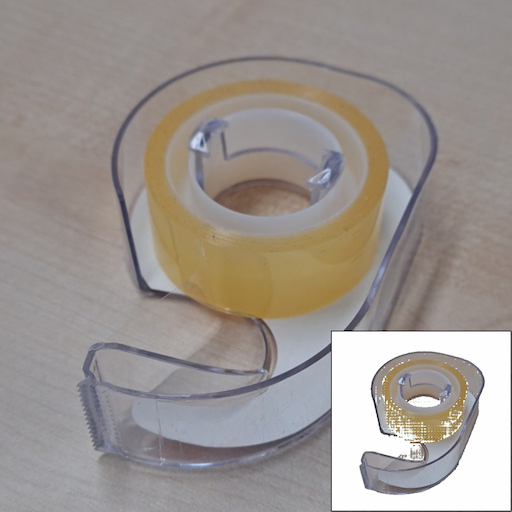}} \\
\midrule
GT & 0.020 & 0.020 & 0.020 & 0.020 & 0.020 \\
\midrule
Ours & \phantom{0}\underline{\tablenum[table-format=1.3]{0.028}} & \bfseries 0.030 & \bfseries 0.022 & \phantom{0}\underline{\tablenum[table-format=1.3]{0.038}} & \underline{0.033} \\
\emph{image2mass} & 0.050 & 0.049 & 0.049 & 0.043 & 0.056 \\
RGB+Depth & 0.059 & \underline{0.031} & 0.036 & \bfseries 0.021 & 0.086 \\
\midrule
\emph{LLaVA (direct)} & 0.001 & 0.001 & 0.001 & 12.300 & 0.001 \\
\emph{LLaVA (reasoning)} & \bfseries 0.015 & 0.850 & 5.000 & 0.100 & 0.094 \\
\emph{Qwen (direct)} & 0.001 & 0.500 & 0.100 & 0.100 & 0.001 \\
\emph{Qwen (reasoning)} & 0.100 & 0.500 & \underline{0.030} & 0.050 & \bfseries 0.010 \\
\bottomrule
\end{tabular}
\caption{Qualitative comparison on household-object images (continued).}
\label{f:real5}
\end{figure}

\begin{figure}[hbt!]
\centering
\setlength{\tabcolsep}{4pt}
\renewcommand{\arraystretch}{1.15}
\begin{tabular}{
    l
    @{\hspace{2pt}}S[table-format=2.3, detect-weight, mode=text]
    @{\hspace{2pt}}S[table-format=2.3, detect-weight, mode=text ]
    @{\hspace{2pt}}S[table-format=2.3, detect-weight, mode=text]
    @{\hspace{2pt}}S[table-format=1.3, detect-weight, mode=text]
    @{\hspace{2pt}}S[table-format=1.3, detect-weight, mode=text]
}
\toprule
&
{\includegraphics[width=0.154\columnwidth]{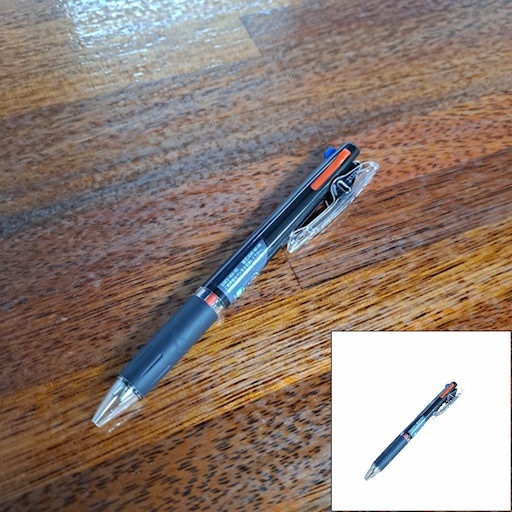}} &
{\includegraphics[width=0.154\columnwidth]{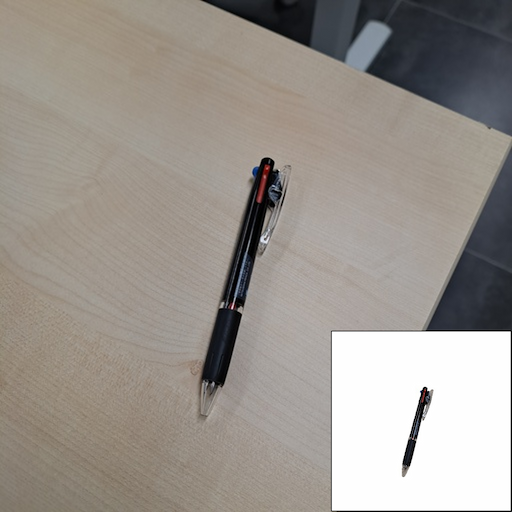}} &
{\includegraphics[width=0.154\columnwidth]{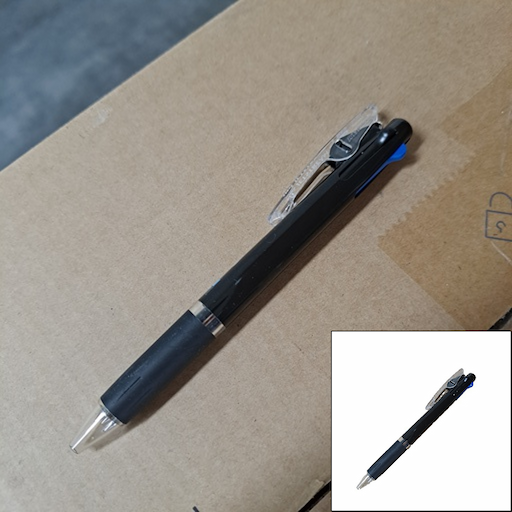}} &
{\includegraphics[width=0.154\columnwidth]{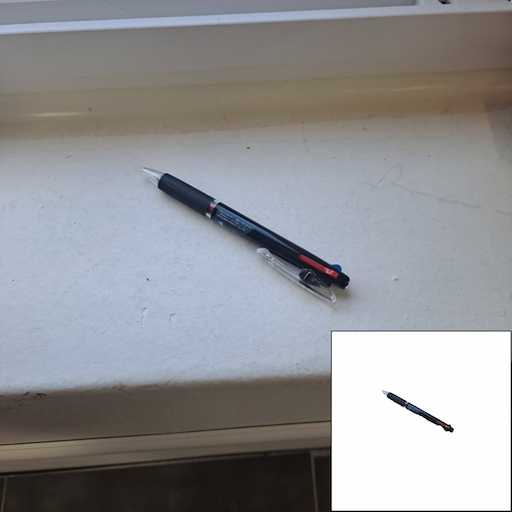}} &
{\includegraphics[width=0.154\columnwidth]{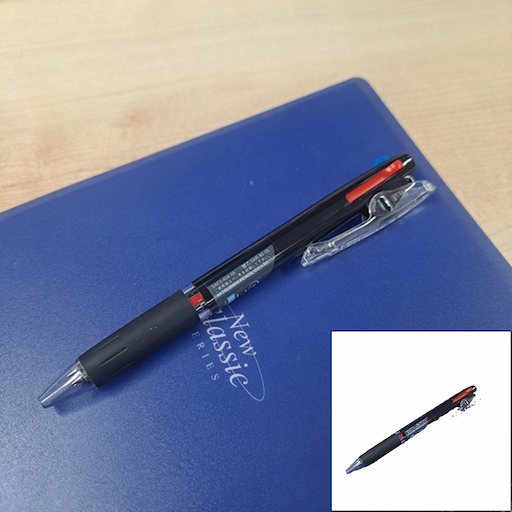}} \\
\midrule
GT & 0.014 & 0.014 & 0.014 & 0.014 & 0.014 \\
\midrule
Ours & \phantom{0}\underline{\tablenum[table-format=1.3]{0.025}} & \phantom{0}\underline{\tablenum[table-format=1.3]{0.018}} & \bfseries 0.018 & 0.022 & 0.026 \\
\emph{image2mass} & 0.034 & 0.060 & 0.037 & 0.126 & 0.045 \\
RGB+Depth & 0.032 & 0.061 & 0.066 & 0.022 & 0.029 \\
\midrule
\emph{LLaVA (direct)} & 0.001 & 0.001 & 0.001 & 0.001 & 0.001 \\
\emph{LLaVA (reasoning)} & 15.000 & \bfseries 0.015 & 15.000 & \textbf{0.015} & \textbf{0.015} \\
\emph{Qwen (direct)} & \bfseries 0.020 & 0.020 & \phantom{0}\underline{\tablenum[table-format=1.3]{0.020}} & 0.050 & \underline{0.020} \\
\emph{Qwen (reasoning)} & \bfseries 0.020 & 0.020 & \phantom{0}\underline{\tablenum[table-format=1.3]{0.020}} & \underline{0.020} & \underline{0.020} \\
\midrule
&
{\includegraphics[width=0.154\columnwidth]{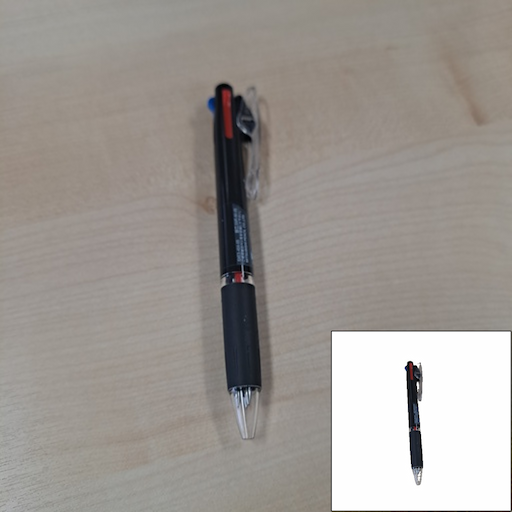}} &
{\includegraphics[width=0.154\columnwidth]{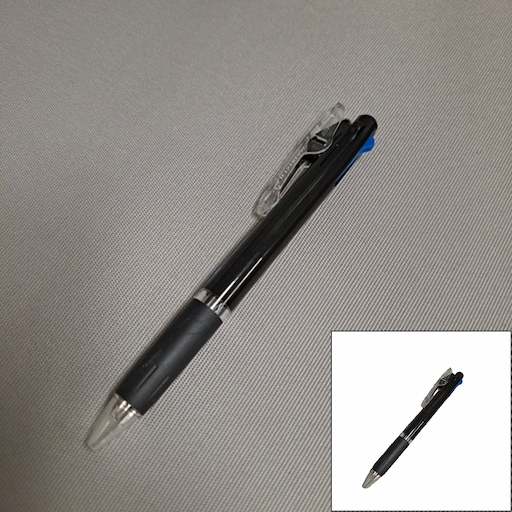}} &
{\includegraphics[width=0.154\columnwidth]{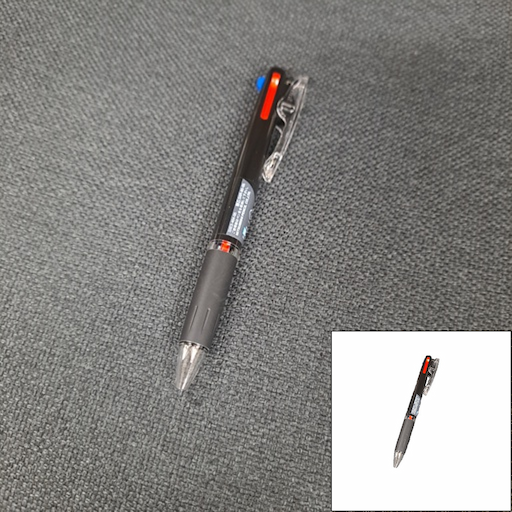}} &
{\includegraphics[width=0.154\columnwidth]{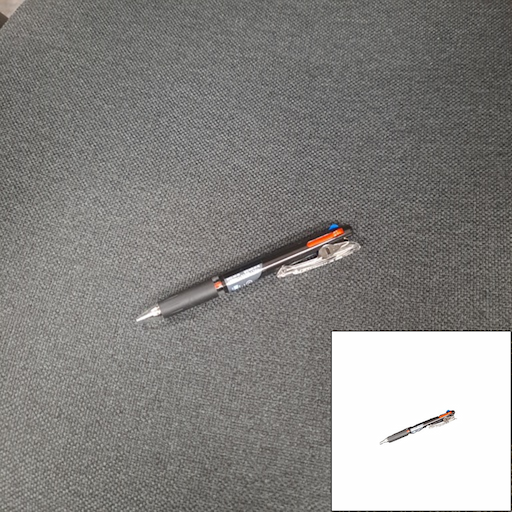}} &
{\includegraphics[width=0.154\columnwidth]{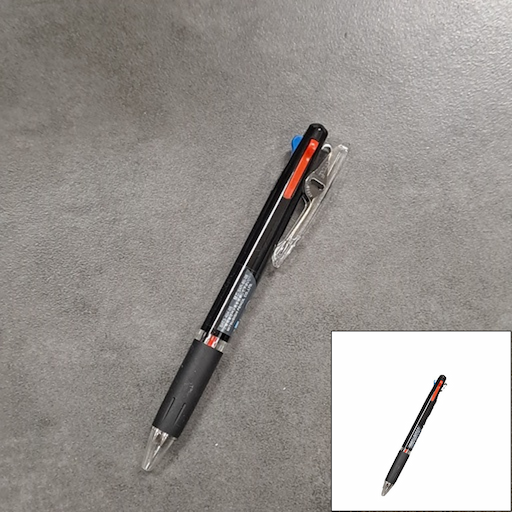}} \\
\midrule
GT & 0.014 & 0.014 & 0.014 & 0.014 & 0.014 \\
\midrule
Ours & 0.028 & \bfseries 0.018 & \phantom{0}\underline{\tablenum[table-format=1.3]{0.017}} & 0.026 & \bfseries 0.017 \\
\emph{image2mass} & 0.056 & 0.038 & 0.040 & 0.071 & 0.065 \\
RGB+Depth & 0.045 & 0.057 & 0.051 & \underline{0.021} & 0.056 \\
\midrule
\emph{LLaVA (direct)} & 0.001 & 0.001 & 0.001 & 0.001 & 0.001 \\
\emph{LLaVA (reasoning)} & 0.100 & 15.000 & 0.100 & 0.100 & 0.100 \\
\emph{Qwen (direct)} & \phantom{0}\underline{\tablenum[table-format=1.3]{0.010}} & \phantom{0}\underline{\tablenum[table-format=1.3]{0.020}} & 0.020 & \bfseries 0.010 & 0.020 \\
\emph{Qwen (reasoning)} & \bfseries 0.015 & 15.000 & \bfseries 0.015 & 0.150 & \underline{0.010} \\
\bottomrule
\end{tabular}
\caption{Qualitative comparison on household-object images (continued).}
\label{f:real6}
\end{figure}

Since our implementation (as well as the compared baselines) assumes object-centric inputs, we preprocess each collected image into a segmented representation to reduce background-induced ambiguity and isolate the target object. We adopt \emph{Lang-SAM},\footnote{\url{https://github.com/luca-medeiros/lang-segment-anything}} a zero-shot text-driven segmentation framework that integrates \emph{GroundingDINO}~\cite{LZR25} and \emph{Segment Anything}~\cite{KMR23}. The overall preprocessing workflow is illustrated in \Cref{f:sam}, and results are shown in \Cref{f:real1,f:real2,f:real3,f:real4,f:real5,f:real6}. Across diverse backgrounds, our approach yields mass estimates closest to the ground truth among single-view baselines.

\section{Limitations and Future Work}
Our approach relies on pre-trained foundation models, including monocular depth predictors for geometry and vision-language models (VLMs) for material semantics. As a result, it may inherit biases from these components, and errors in upstream estimates can propagate through the pipeline.

Since training is supervised only by mass, the decomposition into $\widehat{V}$ and $\hat{\rho}$ is not identifiable up to a multiplicative constant. We therefore interpret them as physically guided latent factors rather than metrically calibrated estimates of volume and density. Recovering calibrated physical quantities may require additional supervision or geometric constraints.

\end{document}